\pdfoutput=1

\documentclass[11pt]{article}

\usepackage{etoolbox}
\def\PaperType{camera}        
\ifdefstring{\PaperType}{camera}{
\usepackage{acl}
}{}
\ifdefstring{\PaperType}{review}{
\usepackage[review]{acl}
}{}

\usepackage{times}
\usepackage{latexsym}

\usepackage[T1]{fontenc}

\usepackage[utf8]{inputenc}

\usepackage{microtype}

\usepackage{bm}
\usepackage{multirow}
\usepackage{amsmath}
\usepackage{amssymb}
\usepackage{mathtools}
\usepackage[inline]{enumitem}
\usepackage{graphicx}
\usepackage{booktabs}
\usepackage{pifont}
\usepackage{mdframed}
\usepackage{dashrule}
\usepackage{caption}
\usepackage{subcaption}
\usepackage{diagbox}
\usepackage{spverbatim}
\usepackage[figuresright]{rotating}
\usepackage{algorithm,algpseudocode}

\algnewcommand{\LineComment}[1]{\State \(\triangleright\) #1}

\newcommand{\matr}[1]{\mathbf{#1}}

\DeclarePairedDelimiterX{\infdivx}[2]{(}{)}{%
  #1\;\delimsize\|\;#2%
}

\renewcommand{\Pr}{\matr{P}}

\newcommand{\weak}[1]{#1\textsubscript{\texttt{wk}}}
\newcommand{\strong}[1]{#1\textsubscript{\texttt{sg}}}
\newcommand{\numStd}[2]{#1\textsubscript{\tiny \textpm #2}}

\title{Lost in Overlap: Exploring Logit-based Watermark Collision in LLMs}

\author{
    Yiyang Luo\textsuperscript{*} \\ {\small Nanyang Technological University} \\ {\ttfamily \small lawrenceluoyy@outlook.com} 
    \And
    Ke Lin\textsuperscript{*} \\ {\small Tsinghua University} \\ {\ttfamily \small leonard.keilin@gmail.com}
    \And
    Chao Gu\textsuperscript{*} \\ {\small University of Science and Technology of China} \\ {\ttfamily \small guch8017@mail.ustc.edu.cn}
    \AND
    Jiahui Hou \\ {\small University of Science and Technology of China} \\ {\ttfamily \small jhhou@ustc.edu.com}
    \And
    Lijie Wen \\ {\small Tsinghua University} \\ {\ttfamily \small wenlj@tsinghua.edu.cn}
    \And
    Ping Luo \\ {\small Tsinghua University} \\ {\ttfamily \small luop@tsinghua.edu.cn}
}

\begin{document}
\maketitle

\ifdefstring{\PaperType}{review}{}{
    \newcommand\freefootnote[1]{%
      \let\thefootnote\relax%
      \footnotetext{#1}%
      \let\thefootnote\svthefootnote%
    }
    \freefootnote{\textsuperscript{*}Equal Contribution.}
    \freefootnote{\textsuperscript{\textdagger}Code and data are available at \url{https://github.com/AInnovateLab/watermark-collision}.}
}

\begin{abstract}
The proliferation of large language models (LLMs) in generating content raises concerns about text copyright. 
Watermarking methods, particularly logit-based approaches, embed imperceptible identifiers into text to address these challenges. 
However, the widespread usage of watermarking across diverse LLMs has led to an inevitable issue known as watermark collision during common tasks, such as paraphrasing or translation.
In this paper, we introduce watermark collision as a novel and general philosophy for watermark attacks, aimed at enhancing attack performance on top of any other attacking methods. 
We also provide a comprehensive demonstration that watermark collision poses a threat to all logit-based watermark algorithms, impacting not only specific attack scenarios but also downstream applications.
\end{abstract}

\section{Introduction}

As the quality of text produced by large language models (LLMs) advances, it addresses numerous practical challenges while raising many new issues. In particular, the widespread generation of text by LLMs on the Internet may increase the spread of rumors and raise concerns about text copyright \cite{megias2021dissimilar, tang2023did}. Consequently, the identification and classification of machine-generated text have become critically significant. 
Watermarking techniques for LLMs can help to tackle these problems, leading to their rising importance in ongoing conversations and attracting increasing interest globally.

Text watermarking involves embedding distinctive, imperceptible identifiers (watermarks) into written content.
Nowadays, most methods are logit-based \cite{kirchenbauer2023KGW, liu2023SIR, zhao2023PRW, kuditipudi2023RDW, hu2023unbiased}: they manipulate the output logits of LLMs during the generation process using distinct but consistently logit-based strategies to embed watermarks successfully.
Utilizing the power of LLMs ensures that such adjustings for probabilistic distribution are seamlessly integrated into the generated content without compromising the overall quality or coherence of the text \cite{lin2024zgls}.
These watermark methods are sophisticatedly designed to be robust yet discreet, ensuring content integrity and ownership preservation without compromising readability or meaning.

\begin{figure}
    \centering
    \includegraphics[width=\linewidth]{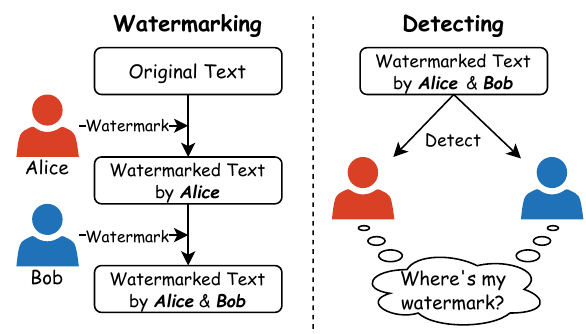}
    \caption{Illustration of watermark collisions.}
    \label{fig:intro}
\end{figure}

However, as watermarking techniques proliferate, watermark collision becomes inevitable with the increasing application of watermarks. 
The term \textbf{watermark~collision} can be defined as instances where the text contains multiple watermarks simultaneously (Fig.~\ref{fig:intro}). This is particularly inevitable during tasks that may require the collaboration of multiple LLMs, such as paraphrasing and translation. While many methods \cite{liu2023SIR, zhao2023PRW, kuditipudi2023RDW} claim resilience against paraphrase attacks, none have been specifically tested for watermark collisions. 
Hence, we examine the underlying mathematical principles, employ watermark collision as an attack strategy for all watermarking techniques that utilize logits, and evaluate its effectiveness in conjunction with multiple traditional attack methods enhanced through the incorporation of watermark collision.

\paragraph{Our Contributions.}
In summary, this paper proposes \textbf{a new watermark attack philosophy for all logit-based watermarks} in LLMs. Our contributions are as follows:
\begin{itemize}
    \item We propose a novel philosophy for watermark attacks that can effectively remove existing watermarks from text. This approach can be integrated with various traditional attack methods to enhance their performance.
    \item We find that the strength of overlapping watermarks impacts detection performance. Upstream and downstream watermarks generally compete for detection accuracy, with one being stronger and the other weaker.
    \item We discuss the vulnerability of watermarking techniques caused by watermark collisions.
\end{itemize}

\section{Related Work}

\paragraph{Text watermarking.}
Modern text watermarking techniques can be classified into two categories: modification-based and generation-based watermarks \cite{liu2024survey}.
Modification-based watermarking, also known as watermarks for existing text, consists of altering an existing text to create a watermarked text. 
Most of the modification-based techniques can be classified as either lexical \cite{topkara2006hiding,yang2022tracing,munyer2023deeptextmark,sato2023embarrassingly} or syntactic methods \cite{atallah2001natural,topkara2006words,meral2009natural}, based on rules, classical machine learning or deep neural models.

\paragraph{LLM watermarking.}
While modification-based techniques \cite{Abdelnabi2020AdversarialWT,yang2022tracing} modify the text and preserve its semantics, generation-based methods apply watermarks into the text generation process to achieve better results, enabling smoother integration with LLMs.
Watermark injection can be carried out during either the training phase or the inference phase.

During the training of LLMs, watermarks are inserted into the training data to intentionally alter the results of LLMs for certain inputs \cite{liu2023watermarking,sun2022coprotector}.
The main objective of training time watermarking is to protect dataset copyrights from unauthorized usage \cite{tang2023did,sun2023codemark}.
Despite being able to embed watermarks in LLMs, training-time watermarking has significant limitations, including limited payload capacity, restricted trigger conditions, and significant training overhead.

For inference-time watermarking, \citet{kirchenbauer2023KGW} proposed a logit-based greenlist mechanism based on prior token hashes.
\citet{liu2023SIR} introduced a watermark model to generate semantic-preserving logits during text generation.
\citet{zhao2023PRW} simplified the \citeauthor{kirchenbauer2023KGW}'s scheme by using a fixed Green-Red split and achieved greater robustness.
\citet{christ2023undetectable,kuditipudi2023RDW,fu2024gumbelsoft} aim to design watermark techniques that are more robust and secure.
\citet{yoo2023advancing,boroujeny2024multi} introduce watermarking techniques with increased payload capacity for arbitrary binary data.
\citet{zhu2024duwak} enhances the efficiency and quality of watermarking by embedding dual secret patterns.

Even though these methods have been designed to be more robust against attacks such as paraphrase attacks \cite{kirchenbauer2023reliability}, back-translation attacks \cite{he2024can} and mask-and-fill attacks \cite{lyu2023adversarial}, these attacks often use unwatermarked LLMs, e.g. DIPPER \cite{krishna2023DIPPER} and GPT-3.5 \cite{brown2020GPT3}. 
Prior research in the pre-LLM era has mentioned potential risks associated with multiple watermarks \cite{tanha2012overview}. Nevertheless, the effects of one watermarking technique on another in the context of LLM watermarking remain unclear, which is the motivation for this study.

\section{Method}

\begin{figure*}[t]
\centering
\scalebox{0.9}{
\includegraphics[width=\linewidth]{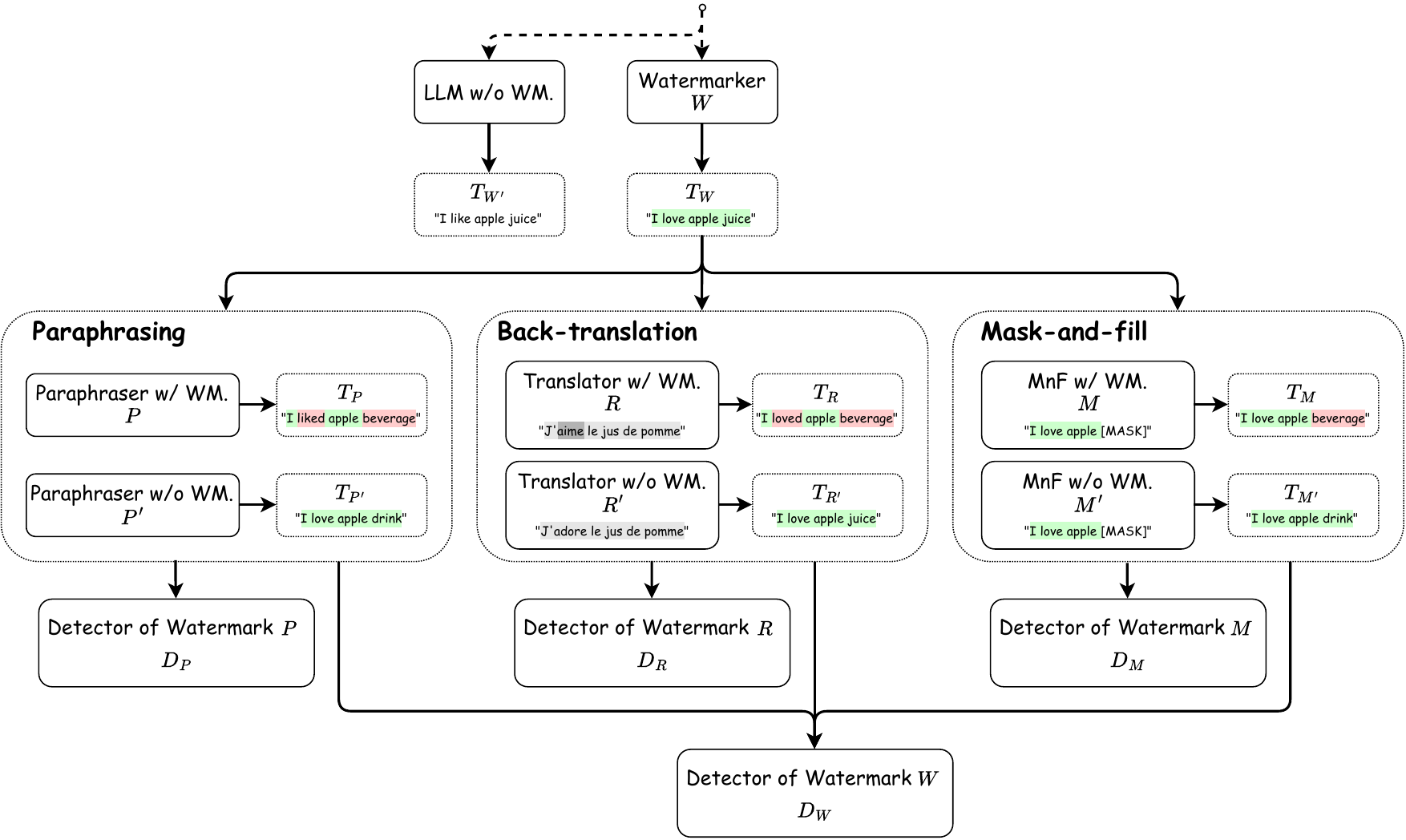}
}
\caption{
The collision pipeline. $T_W$ denotes text with the first watermark $W$, where $T_C$ denotes text with dual watermarks from a different collider $C\in\{P,R,M\}$.
Unwatermarked text generated from $W$ and $C$ is denoted as $T_{W'}$ and $T_{C'}$.
$T_C$ and $T_{C'}$ are then examined by $D_W$ and $D_C$ to determine the presence of watermark $W$ and $C$.
Texts in {\color{red} red} and {\color{green} green} are visualization samples of the red-green list showing the original watermark $W$.
}
\label{fig:pipeline}
\end{figure*}

\subsection{Principle of Watermark Collision}
\label{sec:principle}
The detection process for logit-based watermarking methods relies on the null hypothesis testing. A well-known example is the null hypothesis of KGW \cite{kirchenbauer2023KGW}: 
\begin{equation}
\begin{aligned}
H_0:~&\textit{The text sequence is generated with} \\ 
&\textit{no knowledge of the red list rule.}
\end{aligned}
\end{equation}
Since the words of red list are chosen randomly, a natural writer is expected to sample words both from red and green list, whereas the watermarked model produces words only from green list. In practice, effective detection typically requires the text to contain sufficient words from a runtime-generated green list.
This requirement ensures that the generated text adheres to a specific probability distribution. The detection algorithm checks whether the distribution of words in the text conforms to the green-list distribution. 

Let $\Pr(T)$ represent the probability distribution of a certain text $T$. The detection process verifies if $\Pr(T)$ follows the null hypothesis $H_0$, which can be formulated as the follows, where $\bm{\delta}(\cdot)$ is the dirichlet function:
\begin{equation}
\bm{\delta}[T \text{ is watermarked}]=
\begin{cases}
    0 & \Pr(T) \approx \Pr(H_0) \\
    1 & \Pr(T) \not\approx \Pr(H_0)
\end{cases}
\end{equation}

From a probabilistic perspective, a watermark $w$ can be detected only when text $T$ follows a watermarked distribution $\Pr_w$. 
However, each watermarking method has an independent null hypothesis $H_0$ and therefore creates different word distributions in the generation process.
When an additional watermark $w'$ is applied, it imposes a new distribution $\Pr_{w'}$ on the text. If more watermarks are added, a series of distributions are obtained: $(\Pr_{w^{(0)}}, \Pr_{w^{(1)}}, \Pr_{w^{(2)}}, \dotsc, \Pr_{w^{(n)}})$, where $\Pr_{w^{(0)}}=\Pr_{w}$.
Introducing each new watermark modifies the word probabilities in $T$, thus altering the overall distribution. 
Specifically, any distinct watermarks $w^{(i)}$ and $w^{(j)}$ should have different distributions $\Pr_{w^{(i)}}$ and $\Pr_{w^{(j)}}$ to ensure that they are not incorrectly detected by each other.
As a result, the text $T$ no longer strictly follows the original distribution $\Pr(T)$, nor does it fully conform to any of the subsequent distributions $\Pr_{w^{(0)}}, \Pr_{w^{(1)}}, \Pr_{w^{(2)}}, \dotsc, \Pr_{w^{(n)}}$. Therefore, the combination of multiple watermarks can be described as a transformation of the original watermark into an entangled one:
\begin{equation}
    \Pr_{\text{entangled}} = f(\Pr_{w^{(0)}}, \Pr_{w^{(1)}}, \Pr_{w^{(2)}}, \dotsc, \Pr_{w^{(n)}})
\end{equation}
Here, $f$ represents the complex transformation function resulting from the sequential application of multiple watermarks. This new distribution $\Pr_{\text{entangled}}$ is not merely a simple combination but a new complex distribution that emerges from the (indirect) interaction of all applied watermarks.

Since the detection process relies on identifying patterns consistent with $\Pr_{w^{(i)}}$, the introduction of $\Pr_{\text{entangled}}$ causes the final text distribution to no longer conform to any of these patterns. 
The detection algorithm may not be able to detect the watermark in the text due to a shift in word distribution from $P_{w^{(i)}}$, leading to \emph{watermark collisions}, whereas standard attacks are random, unstable, and less effective.


\subsection{Pipeline Design}
To prove the existence of watermark collisions, we design pipelines with three main components: \emph{watermarker}, \emph{colliders}, and \emph{detectors}. 

\subsubsection{Watermarker}
\emph{Watermarker} $W$ generates watermarked texts $T_W$ by using a language model (LM) to create content based on a specific corpus as context. As illustrated in Fig.~\ref{fig:pipeline}, we first produce the watermarked text data $T_W$ with \emph{Watermarker} $W$. Additionally, we generate unwatermarked text $T_{W'}$ using the same context and prompt as $T_W$ for further comparisons. 
Section~\ref{sec:experiments} and Appendix~\ref{app_sec:pipeline_setup} provide details regarding the watermarker setup.

\subsubsection{Colliders}
\label{sec:colliders}
\emph{Colliders} $C$ are designed to attack the watermark created by the \emph{watermarker} using collision techniques. There are three distinct \emph{colliders} that apply such collision attacks through traditional attack methods, namely paraphraser, back-translator, and mask-and-filler.

\paragraph{Paraphrase Collider.}
\emph{Paraphraser} $P$ rephrases the watermarked texts $T_W$ with different watermarks, i.e. generated by different methods or keys, to generate paraphrased text data $T_P$, which are intended to contain dual watermarks simultaneously. Furthermore, we also generate texts $T_P'$ using the same paraphraser but without a watermark, denoted as $P'$, for further comparison. 

\paragraph{Back-translation Collider.}
\emph{Translator} $R$ translates the watermarked texts $T_W$ to other languages and then translates back to their original language with watermarks. 
As shown in Fig.~\ref{fig:pipeline}, we first paraphrase the original text data $T_W$ using \emph{Watermarker} $W$. Then the text data will be translated and back-translated by \textit{Translator} $R$ with different watermark settings to generate text data $T_R$ which are intended to contain dual watermarks simultaneously.
Furthermore, we also generate texts $T_R'$ using the same translator but without a watermark, denoted as $R'$, for further comparison. 

It is important to note that the KGW-based method is essentially ineffective against back-translation attacks due to the inability to capture the contextual semantics. Thus, they have been excluded from this pipeline. 
We choose French as the pivot language in the back-translation.
Details are in Section~\ref{sec:experiments}.

\paragraph{Mask-and-fill Colliders.}
\emph{Mask-and-filler} (MnF) $M$ is specifically designed for mask-and-fill attacks. The MnF attack method is commonly used with masked language models, e.g., BERT-based models. For our study, we opted for RoBERTa\textsubscript{\texttt{LARGE}} as the base model.
As shown in Fig.~\ref{fig:pipeline}, we first generate watermarked text data $T_W$ using \emph{Watermarker} $W$. Then the text data will be mask-and-filled by \emph{MnF} $M$ with different watermark settings to generate text data $T_M$ which are intended to contain dual watermarks simultaneously.
Additionally, we create texts $T_M'$ by applying the same MnF but excluding the watermark, denoted as $M'$. These will be used as baseline texts for comparison. 

\subsubsection{Detectors}
As demonstrated in Fig.\ref{fig:pipeline}, four detectors are tailored to identify a specific type of watermark. Detector $D_P$ targets watermarks in paraphrasers, $D_R$ focuses on those in translators, and $D_M$ is for watermarks in the MnF process. Detector $D_W$ aims to identify the original watermark embedded by the watermarker. By comparing the results from these detectors, we can assess the effectiveness of the attacks with or without additional watermarks.

\begin{table*}[t]
\renewcommand{\arraystretch}{1.3}
\newcommand{\tableScale}{0.64}
\newcommand{\tableScaleTwo}{0.63}
\begin{subtable}[b]{0.49\textwidth}
\centering
\scalebox{\tableScale}{
\begin{tabular}{c|c|c|cc|cc|cc}
\toprule
\multirow{2}{*}{\diagbox{$W$}{$P$}} & $\varnothing$ & $P'$ & \multicolumn{2}{c|}{\weak{KGW}} & \multicolumn{2}{c|}{\weak{PRW}} & \multicolumn{2}{c}{\weak{SIR}} \\
\cmidrule(lr){2-2} \cmidrule(lr){3-3} \cmidrule(l){4-5} \cmidrule(l){6-7} \cmidrule(l){8-9} & $D_W$ & $D_W$ & $D_W$ & $D_P$ & $D_W$ & $D_P$ & $D_W$ & $D_P$ \\
\midrule
\weak{KGW} & 99.90 & 71.65 & 52.80 & 19.80 & 41.10 & 48.00 & 3.40 & 90.09 \\
\weak{PRW} & 95.40 & 49.25 & 37.00 & 28.20 & 26.60 & 41.20 & 22.30 & 79.56 \\
\weak{SIR} & 87.90 & 59.05 & 55.74 & 25.10 & 41.05 & 19.80 & / & / \\
\bottomrule
\end{tabular}
}
\caption{weak $W$, weak $P$}
\label{tab:P_weak2weak}
\end{subtable}
\hfil
\begin{subtable}[b]{0.49\textwidth}
\centering
\scalebox{\tableScale}{
\begin{tabular}{c|c|c|cc|cc|cc}
\toprule
\multirow{2}{*}{\diagbox{$W$}{$P$}} & $\varnothing$ & $P'$ & \multicolumn{2}{c|}{\weak{KGW}} & \multicolumn{2}{c|}{\weak{PRW}} & \multicolumn{2}{c}{\weak{SIR}} \\
\cmidrule(lr){2-2} \cmidrule(lr){3-3} \cmidrule(l){4-5} \cmidrule(l){6-7} \cmidrule(l){8-9} & $D_W$ & $D_W$ & $D_W$ & $D_P$ & $D_W$ & $D_P$ & $D_W$ & $D_P$ \\
\midrule
\weak{KGW} & 99.90 & 71.65 & 4.10 & 97.40 & 6.20 & 99.90 & 0.20 & 92.77 \\
\weak{PRW} & 95.40 & 49.25 & 14.60 & 96.70 & 9.30 & 99.90 & 29.40 & 91.91 \\
\weak{SIR} & 87.90 & 59.05 & 12.45 & 96.70 & 13.22 & 97.50 & / & / \\
\bottomrule
\end{tabular}
}
\caption{weak $W$, strong $P$}
\label{tab:P_weak2strong}
\end{subtable}

\begin{subtable}[b]{0.49\textwidth}
\centering
\scalebox{\tableScaleTwo}{
\begin{tabular}{c|c|c|cc|cc|cc}
\toprule
\multirow{2}{*}{\diagbox{$W$}{$P$}} & $\varnothing$ & $P'$ & \multicolumn{2}{c|}{\weak{KGW}} & \multicolumn{2}{c|}{\weak{PRW}} & \multicolumn{2}{c}{\weak{SIR}} \\
\cmidrule(lr){2-2} \cmidrule(lr){3-3} \cmidrule(l){4-5} \cmidrule(l){6-7} \cmidrule(l){8-9} & $D_W$ & $D_W$ & $D_W$ & $D_P$ & $D_W$ & $D_P$ & $D_W$ & $D_P$ \\
\midrule
\strong{KGW} & 100.00 & 81.95 & 68.20 & 21.50 & 56.00 & 44.40 & 9.00 & 79.26 \\
\strong{PRW} & 99.70 & 75.05 & 67.90 & 18.30 & 44.20 & 41.70 & 32.00 & 65.31 \\
\strong{SIR} & 94.30 & 67.60 & 61.55 & 21.00 & 45.58 & 14.00 & / & / \\
\bottomrule
\end{tabular}
}
\caption{strong $W$, weak $P$}
\label{tab:P_strong2weak}
\end{subtable}
\hfil
\begin{subtable}[b]{0.49\textwidth}
\centering
\scalebox{\tableScaleTwo}{
\begin{tabular}{c|c|c|cc|cc|cc}
\toprule
\multirow{2}{*}{\diagbox{$W$}{$P$}} & $\varnothing$ & $P'$ & \multicolumn{2}{c|}{\weak{KGW}} & \multicolumn{2}{c|}{\weak{PRW}} & \multicolumn{2}{c}{\weak{SIR}} \\
\cmidrule(lr){2-2} \cmidrule(lr){3-3} \cmidrule(l){4-5} \cmidrule(l){6-7} \cmidrule(l){8-9} & $D_W$ & $D_W$ & $D_W$ & $D_P$ & $D_W$ & $D_P$ & $D_W$ & $D_P$ \\
\midrule
\strong{KGW} & 100.00 & 81.95 & 7.00 & 94.10 & 14.80 & 99.70 & 0.80 & 90.52 \\
\strong{PRW} & 99.70 & 75.05 & 14.60 & 96.70 & 11.20 & 99.30 & 37.10 & 79.00 \\
\strong{SIR} & 94.30 & 67.60 & 10.98 & 97.40 & 12.58 & 97.10 & / & / \\
\bottomrule
\end{tabular}
}
\caption{strong $W$, strong $P$}
\label{tab:P_strong2strong}
\end{subtable}

\caption{TPR of the \emph{paraphrased} text $T_P$ with dual watermarks when $\text{FPR}=1\%$. 
$W$ and $P$ represent the watermarker and paraphraser, respectively. 
$D_W$ and $D_P$ represent the detector of the watermarker and paraphraser. $\varnothing$ indicates that no paraphrasing process is applied to the text, and its corresponding column represents the result of using $D_W$ to detect watermark $W$ in $T_W$. 
$P'$ represents paraphrasing $T_W$ without watermark, as mentioned in Fig.~\ref{fig:pipeline}.}
\label{tab:P_FPR_TPR}
\end{table*}


\section{Experiments}
\label{sec:experiments}

\subsection{Experiment Setup}
\label{sec:experiment_setup}
\paragraph{Settings.}
We utilize the C4 dataset \cite{raffel2020exploring} as the context for text generation with a maximum of 128 tokens. 
For watermarker and paraphraser, we employ LLaMA-2-13B \cite{touvron2023llama2}, Qwen2-7B \cite{bai2023qwen} and OPT-1.3B \cite{zhang2022opt}. 
For back-translator, we use LLaMA-2-13B and Qwen2-7B.
For mask-and-filler, we exclusively utilize RoBERTa\textsubscript{\texttt{LARGE}} \cite{liu2019roberta} as the masked language model.
We only present the results of LLaMA-2-13B here and refer to the Appendix~\ref{app_sec:experimental_results} for more details.

\paragraph{Watermarks.}

Our experiments were conducted using several previous watermark methods as baselines. The watermark strength of each method is tuned separately for \texttt{weak} and \texttt{strong} watermarks. Methods involved include:
\begin{enumerate*}[label=(\alph*)]
    \item \textbf{KGW} from \citet{kirchenbauer2023KGW}, $d=2,5$ for weak and strong settings;
    \item \textbf{PRW} from \citet{zhao2023PRW}, $s=2,5$ for weak and strong settings.
    \item \textbf{SIR} from \citet{liu2023SIR}, $d=2,5$ for weak and strong settings;
\end{enumerate*}
Note that we use subscript \weak{$\cdot$} and \strong{$\cdot$} to refer to weak and strong watermarks, respectively.

Some methods are not selected for specific reasons.
\textbf{UBW} from \citet{hu2023unbiased} asserting that their approach lacks resilience against paraphrase attacks, thereby rendering it unsuitable for this experiment.
\textbf{RDW} from \citet{kuditipudi2023RDW} may be ineffective when subjected to a key different from their recommended configuration, thereby failing to meet the experimental requirements that necessitate distinct key settings.

\subsection{Evaluation Metrics}
Given the rapid development of the watermarking field, there is currently no consensus on metrics, with different methods employing varied evaluation criteria \cite{tu2023waterbench}. Moreover, some approaches utilize a detection threshold to categorize text as watermarked, while others differ. Implementing all possible metrics for each method is impractical and biased. 

Following the approach of \citet{zhao2023PRW}, we opt for a fair evaluation by avoiding the influence of threshold settings. Our final detection metrics include false positive rates (\textbf{FPR}) and true positive rates (\textbf{TPR}). 
We specifically set FPR values at 1\%, 5\%, and 10\%, adjusting the detector's thresholds accordingly. This ensures a consistent and unbiased assessment of watermarking methods.
We only present results when $\text{FPR}=1\%$ here and refer to Appendix~\ref{app_sec:experimental_results} for comprehensive results.

\begin{table*}
\renewcommand{\arraystretch}{1.3}
\newcommand{\tableScale}{0.64}
\newcommand{\tableScaleTwo}{0.63}
\begin{subtable}[b]{0.49\textwidth}
\centering
\scalebox{\tableScale}{
\begin{tabular}{c|c|c|cc|cc|cc}
\toprule
\multirow{2}{*}{\diagbox{$W$}{$R$}} & $\varnothing$ & $R'$ & \multicolumn{2}{c|}{\weak{KGW}} & \multicolumn{2}{c|}{\weak{PRW}} & \multicolumn{2}{c}{\weak{SIR}} \\
\cmidrule(lr){2-2} \cmidrule(lr){3-3} \cmidrule(l){4-5} \cmidrule(l){6-7} \cmidrule(l){8-9} & $D_W$ & $D_W$ & $D_W$ & $D_R$ & $D_W$ & $D_R$ & $D_W$ & $D_R$ \\
\midrule
\weak{KGW} & 99.90 & 44.80 & 41.90 & 9.20 & 34.20 & 13.70 & 31.40 & 5.15 \\
\weak{PRW} & 95.40 & 32.90 & \underline{34.10} & 7.50 & 21.80 & 12.70 & 25.20 & 5.73 \\
\weak{SIR} & 87.90 & 5.60 & 4.75 & 7.70 & \underline{6.12} & 11.50 & / & / \\
\bottomrule
\end{tabular}
}
\caption{weak $W$, weak $R$}
\label{tab:R_weak2weak}
\end{subtable}
\hfil
\begin{subtable}[b]{0.49\textwidth}
\centering
\scalebox{\tableScale}{
\begin{tabular}{c|c|c|cc|cc|cc}
\toprule
\multirow{2}{*}{\diagbox{$W$}{$R$}} & $\varnothing$ & $R'$ & \multicolumn{2}{c|}{\weak{KGW}} & \multicolumn{2}{c|}{\weak{PRW}} & \multicolumn{2}{c}{\weak{SIR}} \\
\cmidrule(lr){2-2} \cmidrule(lr){3-3} \cmidrule(l){4-5} \cmidrule(l){6-7} \cmidrule(l){8-9} & $D_W$ & $D_W$ & $D_W$ & $D_R$ & $D_W$ & $D_R$ & $D_W$ & $D_R$ \\
\midrule
\weak{KGW} & 99.90 & 44.80 & 28.50 & 69.50 & 18.50 & 58.50 & 3.90 & 93.50 \\
\weak{PRW} & 95.40 & 32.90 & 23.00 & 68.00 & 14.30 & 54.60 & 10.80 & 92.87 \\
\weak{SIR} & 87.90 & 5.60 & 4.28 & 67.40 & 4.40 & 41.60 & / & / \\
\bottomrule
\end{tabular}
}
\caption{weak $W$, strong $R$}
\label{tab:R_weak2strong}
\end{subtable}

\begin{subtable}[b]{0.49\textwidth}
\centering
\scalebox{\tableScaleTwo}{
\begin{tabular}{c|c|c|cc|cc|cc}
\toprule
\multirow{2}{*}{\diagbox{$W$}{$R$}} & $\varnothing$ & $R'$ & \multicolumn{2}{c|}{\weak{KGW}} & \multicolumn{2}{c|}{\weak{PRW}} & \multicolumn{2}{c}{\weak{SIR}} \\
\cmidrule(lr){2-2} \cmidrule(lr){3-3} \cmidrule(l){4-5} \cmidrule(l){6-7} \cmidrule(l){8-9} & $D_W$ & $D_W$ & $D_W$ & $D_R$ & $D_W$ & $D_R$ & $D_W$ & $D_R$ \\
\midrule
\strong{KGW} & 100.00 & 69.30 & 67.80 & 8.10 & 61.90 & 13.60 & 55.30 & 4.09 \\
\strong{PRW} & 99.70 & 63.90 & \underline{64.80} & 9.60 & 55.20 & 10.20 & 50.60 & 3.56 \\
\strong{SIR} & 94.30 & 3.00 & 1.74 & 9.40 & \underline{3.82} & 6.70 & / & / \\
\bottomrule
\end{tabular}
}
\caption{strong $W$, weak $R$}
\label{tab:R_strong2weak}
\end{subtable}
\hfil
\begin{subtable}[b]{0.49\textwidth}
\centering
\scalebox{\tableScaleTwo}{
\begin{tabular}{c|c|c|cc|cc|cc}
\toprule
\multirow{2}{*}{\diagbox{$W$}{$R$}} & $\varnothing$ & $R'$ & \multicolumn{2}{c|}{\weak{KGW}} & \multicolumn{2}{c|}{\weak{PRW}} & \multicolumn{2}{c}{\weak{SIR}} \\
\cmidrule(lr){2-2} \cmidrule(lr){3-3} \cmidrule(l){4-5} \cmidrule(l){6-7} \cmidrule(l){8-9} & $D_W$ & $D_W$ & $D_W$ & $D_R$ & $D_W$ & $D_R$ & $D_W$ & $D_R$ \\
\midrule
\strong{KGW} & 100.00 & 69.30 & 50.00 & 68.90 & 38.80 & 53.00 & 7.50 & 91.70 \\
\strong{PRW} & 99.70 & 63.90 & 50.10 & 67.90 & 36.60 & 42.40 & 20.10 & 90.44 \\
\strong{SIR} & 94.30 & 3.00 & \underline{4.15} & 74.40 & 2.51 & 22.50 & / & / \\
\bottomrule
\end{tabular}
}
\caption{strong $W$, strong $R$}
\label{tab:R_strong2strong}
\end{subtable}

\caption{TPR of the \emph{back-translated} text $T_R$ with dual watermarks when $\text{FPR}=1\%$. Similar to Table~\ref{tab:P_FPR_TPR}.
Data annotated with \underline{underline} indicate abnormal data points.
}
\label{tab:R_FPR_TPR}
\end{table*}


\begin{table}[t]
\renewcommand{\arraystretch}{1.3}
\newcommand{\tableScale}{0.8}
\newcommand{\tableScaleTwo}{0.63}
\centering
\begin{subtable}[b]{0.49\textwidth}
\centering
\scalebox{\tableScale}{
\begin{tabular}{c|c|c|c|c|c}
\toprule
\diagbox{$W$}{$P$} & $\varnothing$ & $P'$ & \weak{KGW} & \weak{PRW} & \weak{SIR} \\
\midrule
\weak{KGW} & 7.36 & 6.96 & 8.60 & 6.58 & 12.28 \\
\weak{PRW} & 6.92 & 6.69 & 9.77 & 6.24 & 12.15 \\
\weak{SIR} & 10.14 & 7.15 & 11.04 & 7.09 & / \\
\bottomrule
\end{tabular}
}
\caption{weak $W$, weak $P$}
\label{tab:ppl_weak2weak}
\end{subtable}
\hfil
\begin{subtable}[b]{0.49\textwidth}
\centering
\scalebox{\tableScale}{
\begin{tabular}{c|c|c|c|c|c}
\toprule
\diagbox{$W$}{$P$} & $\varnothing$ & $P'$ & \strong{KGW} & \strong{PRW} & \strong{SIR} \\
\midrule
\weak{KGW} & 7.36 & 6.96 & 15.67 & 5.05 & 14.08 \\
\weak{PRW} & 6.92 & 6.69 & 13.85 & 5.22 & 10.25 \\
\weak{SIR} & 10.14 & 7.15 & 13.31 & 5.12 & / \\
\bottomrule
\end{tabular}
}
\caption{weak $W$, strong $P$}
\label{tab:ppl_weak2strong}
\end{subtable}

\begin{subtable}[b]{0.49\textwidth}
\centering
\scalebox{\tableScale}{
\begin{tabular}{c|c|c|c|c|c}
\toprule
\diagbox{$W$}{$P$} & $\varnothing$ & $P'$ & \weak{KGW} & \weak{PRW} & \weak{SIR} \\
\midrule
\strong{KGW} & 13.18 & 7.44 & 14.16 & 8.39 & 13.55 \\
\strong{PRW} & 8.31 & 7.61 & 10.53 & 6.54 & 12.00 \\
\strong{SIR} & 12.30 & 7.72 & 12.75 & 7.80 & / \\
\bottomrule
\end{tabular}
}
\caption{strong $W$, weak $P$}
\label{tab:ppl_strong2weak}
\end{subtable}
\hfil
\begin{subtable}[b]{0.49\textwidth}
\centering
\scalebox{\tableScale}{
\begin{tabular}{c|c|c|c|c|c}
\toprule
\diagbox{$W$}{$P$} & $\varnothing$ & $P'$ & \strong{KGW} & \strong{PRW} & \strong{SIR} \\
\midrule
\strong{KGW} & 13.18 & 7.44 & 12.80 & 5.60 & 12.03 \\
\strong{PRW} & 8.31 & 7.61 & 11.15 & 4.91 & 9.22 \\
\strong{SIR} & 12.30 & 7.72 & 11.29 & 4.99 & / \\
\bottomrule
\end{tabular}
}
\caption{strong $W$, strong $P$}
\label{tab:ppl_strong2strong}
\end{subtable}

\caption{PPL of the \textit{paraphrased} $T_P$ with dual WMs.}
\label{tab:ppl_p}
\end{table}


\begin{table}[t]
\renewcommand{\arraystretch}{1.3}
\newcommand{\tableScale}{0.8}
\newcommand{\tableScaleTwo}{0.63}
\centering
\begin{subtable}[b]{0.49\textwidth}
\centering
\scalebox{\tableScale}{
\begin{tabular}{c|c|c|c|c|c}
\toprule
\diagbox{$W$}{$P$} & $\varnothing$ & $P'$ & \weak{KGW} & \weak{PRW} & \weak{SIR} \\
\midrule
\weak{KGW} & 7.36 & 8.82 & 9.00 & 8.29 & 10.16 \\
\weak{PRW} & 6.92 & 8.21 & 8.52 & 8.02 & 9.39 \\
\weak{SIR} & 10.14 & 9.82 & 10.37 & 9.14 & / \\
\bottomrule
\end{tabular}
}
\caption{weak $W$, weak $R$}
\label{tab:ppl_bt_weak2weak}
\end{subtable}
\hfil
\begin{subtable}[b]{0.49\textwidth}
\centering
\scalebox{\tableScale}{
\begin{tabular}{c|c|c|c|c|c}
\toprule
\diagbox{$W$}{$P$} & $\varnothing$ & $P'$ & \strong{KGW} & \strong{PRW} & \strong{SIR} \\
\midrule
\weak{KGW} & 7.36 & 8.82 & 11.86 & 9.63 & 33.35 \\
\weak{PRW} & 6.92 & 8.21 & 11.15 & 8.84 & 30.57 \\
\weak{SIR} & 10.14 & 9.82 & 13.39 & 9.92 & / \\
\bottomrule
\end{tabular}
}
\caption{weak $W$, strong $R$}
\label{tab:ppl_bt_weak2strong}
\end{subtable}

\begin{subtable}[b]{0.49\textwidth}
\centering
\scalebox{\tableScale}{
\begin{tabular}{c|c|c|c|c|c}
\toprule
\diagbox{$W$}{$P$} & $\varnothing$ & $P'$ & \weak{KGW} & \weak{PRW} & \weak{SIR} \\
\midrule
\strong{KGW} & 13.18 & 12.74 & 12.88 & 11.41 & 14.91 \\
\strong{PRW} & 8.31 & 9.02 & 9.29 & 8.33 & 11.11 \\
\strong{SIR} & 12.30 & 10.05 & 10.91 & 8.72 & / \\
\bottomrule
\end{tabular}
}
\caption{strong $W$, weak $R$}
\label{tab:ppl_bt_strong2weak}
\end{subtable}
\hfil
\begin{subtable}[b]{0.49\textwidth}
\centering
\scalebox{\tableScale}{
\begin{tabular}{c|c|c|c|c|c}
\toprule
\diagbox{$W$}{$P$} & $\varnothing$ & $P'$ & \strong{KGW} & \strong{PRW} & \strong{SIR} \\
\midrule
\strong{KGW} & 13.18 & 12.74 & 17.58 & 11.83 & 36.84 \\
\strong{PRW} & 8.31 & 9.02 & 12.59 & 9.60 & 27.98 \\
\strong{SIR} & 12.30 & 10.05 & 14.78 & 9.18 & / \\
\bottomrule
\end{tabular}
}
\caption{strong $W$, strong $R$}
\label{tab:ppl_bt_strong2strong}
\end{subtable}

\caption{PPL of the \textit{back-translated} $T_R$ with dual WMs.}
\label{tab:ppl_bt}
\end{table}

\subsection{Experimental Results \& Analysis}

\paragraph{Watermark collision is compatible with the majority of existing attacks.}
Tables \ref{tab:P_FPR_TPR}, \ref{tab:R_FPR_TPR}, and \ref{tab:M_Z_Score} demonstrate that watermark collision is feasible for all selected attacks, including paraphrase, back-translation, and mask-and-fill attacks.
Watermark collision is commonly found in attacks involving auto-regressive text generation methods such as paraphrasing and back-translation. Mask-and-fill attacks are ineffective as they cannot completely change the distribution of words in a sentence. 

\paragraph{Watermark collision will not degrade the text quality.}
As shown in Tab. \ref{tab:ppl_p} and \ref{tab:ppl_bt}, we present perplexity before and after attacks, both with and without collisions.
We use LLaMA-2-13B as the backbone for perplexity calculation, which is the same model for the initial generated text. 
As evidence, text quality remains largely stable post-attack, and most collisions did not result in significant declines in text quality, indicating the potential value of collision as an attack methodology. 
Detailed semantic analysis examples of the text quality can be found in Appendix~\ref{app_sec:experimental_results}.

\paragraph{Watermark attacks with watermarks collision tend to be stronger than those without.}
In Table~\ref{tab:P_FPR_TPR} and \ref{tab:R_FPR_TPR}, we present the detection accuracy for various baseline watermark algorithms with and without the occurrence of watermark collision ($T_P$ and $T_P'$, respectively). A noteworthy decline in detection accuracy is observed when watermarks are introduced in the context of traditional attacks such as paraphrase attacks and back-translation attacks.
According to the settings in Table~\ref{tab:P_FPR_TPR} and \ref{tab:R_FPR_TPR}, there is a strong competition between overlapping watermarks. As one watermarker attempts to maintain its detection accuracy, the others' detection accuracy decreases. 
SIR-SIR is not listed since the SIR does not allow the user to choose a key to create a different distribution, and therefore the watermark collision will not occur on SIR-SIR. However, SIR watermarks are still vulnerable to collision attacks using other watermark methods.

It should also be noted that different watermarking methods behave differently during competition. 
KGW appears to be less competent than the other two methods. SIR, however, shows significant collisions even in weak paraphraser settings (Column SIR of Tab.~\ref{tab:P_weak2weak} \& \ref{tab:P_strong2weak}), while PRW exhibits extreme collisions in strong paraphraser settings (Column PRW of Tab.~\ref{tab:P_weak2strong} \& \ref{tab:P_strong2strong}).

However, some anomalies can be observed in back-translation (Table \ref{tab:R_FPR_TPR}), where the TPR increases after a collision. These anomalies occur because certain watermarks may have similar distributions in specific contexts, rendering them ineffective and making attacks on them pointless:  
When two watermark methods exhibit similar distributions, the likelihood of collision decreases, leading to an increase in the True Positive Rate (TPR). 
However, in fact, this is a degraded performance for a certain watermark method: It is a challenge to identify the origin of the watermark. In practical scenarios, if the origin of the watermark cannot be determined, it essentially means we cannot ascertain which entity embeds the watermark and makes the watermark meaningless. 
For example, if two KGW-based watermark methods utilize the \textit{a similar but not the same} red-green list, paraphrasing words into other similar words may accidentally reinforce the watermark within the text, thereby enhancing the TPR for both watermarks and lead to confusion on which entity apply watermark on the text.
Consequently, such cases are considered \textbf{failures} of watermark methods and do not concern us.


\paragraph{MnF exhibits a similar trend as the watermark collision intensifies, however, it experiences less impact compared to other colliders.}
We note that attacks on MnF are less effective because the unmasked words retain most of the context and keep the watermark unaltered. Therefore, the TPR in these cases remains nearly the same. 

Nonetheless, it is observed that when considering the z-score of $H_0$ as stated in \cite{kirchenbauer2023KGW}, the influence of watermark collision persists.
In Table~\ref{tab:M_Z_Score}, we present the z-scores for each corresponding method. 
Degradation of the z-score is observed after collision attacks compared to those without, indicating the effectiveness of collisions. 
Besides, the mask rate has a greater impact on the detection process.
If the mask rate is low, MnF is less likely to have a significant effect, thus explaining why these attacks have minimal impact on MnF.
Furthermore, the SIR method is excluded from MnF experiments, as it prioritizes semantic factors and is less susceptible to MnF attacks.

\begin{table}
\renewcommand{\arraystretch}{1.15}
\newcommand{\tableScale}{0.76}
\centering
\scalebox{\tableScale}{
\begin{tabular}{c|ccccc}
\toprule
\diagbox{$W$}{$M$} & $\varnothing$ & \weak{KGW} & \weak{PRW} & \strong{KGW} & \strong{PRW} \\
\midrule
\weak{KGW} & \numStd{4.14}{1.57} & \numStd{3.88}{1.51} & \numStd{3.62}{1.52} & \numStd{3.52}{1.51} & \numStd{3.42}{1.56} \\
\weak{PRW} & \numStd{5.14}{1.51} & \numStd{4.89}{1.52} & \numStd{4.64}{1.59} & \numStd{4.35}{1.60} & \numStd{4.44}{1.57} \\
\strong{KGW} & \numStd{6.72}{1.95} & \numStd{6.37}{1.95} & \numStd{6.09}{1.99} & \numStd{5.97}{1.87} & \numStd{5.86}{1.95} \\
\strong{PRW} & \numStd{7.82}{1.70} & \numStd{7.62}{1.74} & \numStd{7.35}{1.81} & \numStd{7.08}{1.87} & \numStd{7.07}{1.82} \\
\bottomrule
\end{tabular}
}
\caption{
Z-scores of the \emph{mask-and-filled} text $T_M$ with dual watermarks with a mask rate of $0.6$ for masked language modeling tasks. 
$\varnothing$ means no MnF is applied, i.e., the original detection results of z-scores.
}
\label{tab:M_Z_Score}
\end{table}

\subsection{Multi-round Collision}

\begin{figure*}[t]
\newcommand{\imageSize}{0.45\linewidth}
\centering
\begin{subfigure}[t]{\imageSize}
    \centering
    \includegraphics[width=\linewidth]{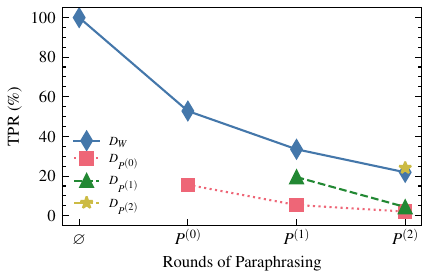}
    \caption{$P=\text{\weak{KGW}}$}
\end{subfigure}
\begin{subfigure}[t]{\imageSize}
    \centering
    \includegraphics[width=\linewidth]{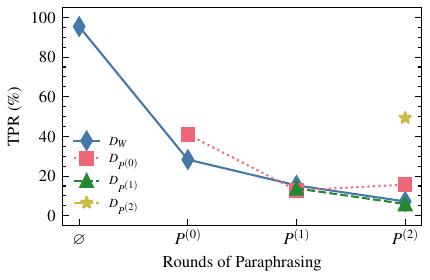}
    \caption{$P=\text{\weak{PRW}}$}
\end{subfigure}
\begin{subfigure}[t]{\imageSize}
    \centering
    \includegraphics[width=\linewidth]{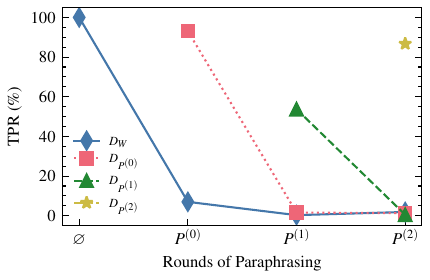}
    \caption{$P=\text{\strong{KGW}}$}
\end{subfigure}
\begin{subfigure}[t]{\imageSize}
    \centering
    \includegraphics[width=\linewidth]{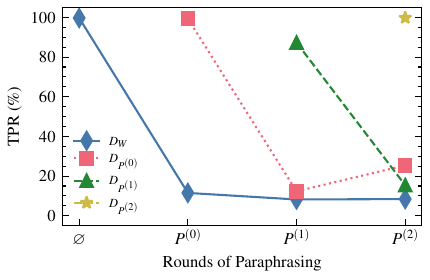}
    \caption{$P=\text{\strong{PRW}}$}
\end{subfigure}
\caption{ 
Multi-round TPR of \emph{paraphrased} text under a series of paraphrase attacks by the same type of paraphraser with different watermarks.
$\varnothing$ represents the original detection TPR before paraphrasing. A sequence of paraphrasers $(P^{(0)},P^{(1)},P^{(2)},\dotsc)$ is applied consecutively to the generated text from the preceding paraphraser.
}
\label{fig:multi_round}
\end{figure*}

To further enhance the performance of these attacks, multi-round collisions can be applied. 
We tested the performance of multi-round attacks both with and without watermark collisions, and the results are presented in Fig.~\ref{fig:multi_round}. 
In each experiment, we use the same paraphraser to assess the chain effect caused by watermark collisions.
For various watermark methods, the TPR of each watermark detection decreases after multi-round collision attacks. 
As explained in the previous section, 
applying multi-round collisions causes the word distribution to deviate more significantly from subsequent distributions.
The stronger the watermark, the less likely it is for the multi-round watermark to coexist with others.


\section{Possible Application}

\paragraph{Malicious attacks based on watermark collisions.}
Previous works \cite{kirchenbauer2023reliability, kirchenbauer2023KGW, kuditipudi2023RDW} have introduced several attacks, such as copy-paste attacks and paraphrase attacks, but most have shown their robustness and security against at least some of these attacks. 
Our study, however, provides a feasible method of constructing effective attacks using watermark collisions. 
For example, text with a \weak{KGW} watermark can be detected with a 44.8\% TPR after a paraphrase attack without a watermark. However, if the paraphrase attack is conducted with another \weak{KGW}, the detection TPR drops to 41.9\%. If the attack is done with a \strong{KGW}, the detection TPR further decreases to 28.5\%, as presented in Table \ref{tab:P_weak2weak} and \ref{tab:P_weak2strong}. 
The use of colliders with strong watermarks could easily erase existing watermarks, resulting in greater vulnerability to watermarking. 

\paragraph{Detection of existing watermarks using collisions between watermarks of different strengths.}
In Table~\ref{tab:P_FPR_TPR} and \ref{tab:R_FPR_TPR}, we demonstrate that weak watermarks can still be easily applied to unwatermarked text. It is, however, much more difficult to apply a weak watermark to text that has already been watermarked (Tab.~\ref{tab:P_FPR_TPR} \& \ref{tab:R_FPR_TPR}). 
For example, KGW can be applied to plaintext with a success rate of 99.90\%, but can be applied to a SIR-watermarked text with a probability not exceeding 25.10\%, as presented in Table \ref{tab:P_weak2weak} and \ref{tab:P_strong2weak}.
This provides a simple probabilistic method of detecting watermarks without the need to know their details. When adding a weak watermark to a sentence is difficult, it is more likely to have an existing one in the original sentence. 

\section{Discussion}
The LLM watermarking technique is currently undergoing rapid development, with many foundational aspects not yet implemented for practical use regularly. 
However, our research demonstrates that when watermarks collide, it can significantly hinder the performance of the watermark when applied in real-world situations.
A list of predictable risks in practical applications is provided:
\begin{itemize}
    \item \textbf{API Tracing}: LLM providers can use watermarking techniques on their LLM API to prevent unauthorized use. However, if the watermarked output of LLMs is sent to other providers with watermarks for further processing, the upstream watermarks will not be effective in tracing the use of the upstream APIs.
    \item \textbf{Black-box Detection}: As we discussed in the Experiments Section, the watermark collisions could perform black-box detection of any existing watermarks in any text. Users would experience distrust when they become aware of the presence of watermarks. Hackers may attempt to bypass the watermarks.
\end{itemize}

\section{Conclusion}
In this study, we examine how overlapping watermarks in the same text can decrease the accuracy of both upstream and downstream watermark detection. 
We propose the use of \textbf{watermark collisions as an attacking philosophy} and therefore emphasize that watermark collisions may compromise the validity and security of all logit-based watermarks. 
We conduct experiments by integrating various watermarkers and colliders, assessing the text quality before and after collisions, as well as with and without collisions, to demonstrate that collision can enhance common attacks and text quality remains consistent when compared to attacks without collisions.
We hope our work will increase awareness of potential threats to LLM watermarking.

\section*{Limitations}
Our approach indeed demonstrates the potential collision between existing watermark techniques. Nevertheless, we conduct our experiments only on paraphrasing, back-translation, and mask-and-fill tasks to simulate watermark collisions. Broader but less related tasks, such as question answering, have not yet been tested.

Furthermore, the range of models chosen is limited. The models chosen include LLaMA-2-13B, Qwen2-7B, OPT-1.3B, and RoBERTa\textsubscript{\texttt{LARGE}} as the colliders, while certain other models are tailored for particular tasks. Experiments on more open-sourced models could further enhance the conclusion of our paper.

%


\bibliography{IEEEabrv,main}

\appendix

\newpage
\section*{Appendix}
\section{Pipeline Setup}
\label{app_sec:pipeline_setup}
\paragraph{Datasets.}
To generate the watermarked text $T_W$, the C4 dataset is used as the context. A total of 1000 watermarked textual samples are generated from the selected context. 
To ensure that only text with an apparent watermark is selected for $T_W$, a z-score threshold is set during the generation.
For KGW and PRW, the z-threshold is set to $4.0$. For SIR, the z-threshold is set to $0.0$.

\paragraph{Hyperparameters.}
We specifically designate $2024$ as the key for the watermarker (if applicable) and $2023$ as the key for the paraphraser (if applicable) to vary the key for watermark collision within the same watermark algorithm (e.g. \textbf{KGW}-\textbf{KGW}).
The specific hyperparameters of each watermarking method are as follows:
\begin{itemize}
    \item \textbf{KGW}: For weak settings, we set the green list size $\gamma=0.25$, hardness parameter $\delta=2.0$ and the seeding scheme is \texttt{selfhash}. For strong settings, we set the green list size $\gamma=0.25$, hardness parameter $\delta=5.0$ and the seeding scheme is \texttt{selfhash}.
    \item \textbf{SIR}: We employ the \texttt{context} mode. For weak settings, We set $\text{chunk\_length}=10$, $\gamma=0.5$, watermark strenght $\delta=2.0$. For strong settings, we set $\text{chunk\_length}=10$, $\gamma=0.5$, watermark strength $\delta=5.0$.
    \item \textbf{PRW}: For weak settings, we set the green list size $\gamma=0.25$, watermark strength $\delta=2.0$. For strong settings, we set the green list size $\gamma=0.25$, watermark strength $\delta=5.0$.
\end{itemize}

\begin{figure}[hb]
\small
\ttfamily
\begin{mdframed}
<<SYS>>
\newline
Assume you are a helpful assistant. Your job is to paraphrase the given text.
\newline
<</SYS>>
\newline
\newline
[INST]\texttt{\{INPUT\_TEXT\}}[/INST]
\newline
\newline
You're welcome! Here's a paraphrased version of the original message:
\end{mdframed}
\caption{The paraphrase prompt template for LLaMA-2 paraphraser.}
\label{fig:prompts}
\end{figure}

\begin{figure}[hb]
\small
\ttfamily
\begin{mdframed}
<|im\_start|>system
\newline
You are a helpful assistant. Your job is to paraphrase the given text.<|im\_end|>\newline
<|im\_start|>user\newline
\texttt{\{INPUT\_TEXT\}}<|im\_end|>\newline
<|im\_start|>assistant\newline
You're welcome! Here's a paraphrased version of the original message: 
\end{mdframed}
\caption{The paraphrase prompt template for Qwen2 paraphraser.}
\label{fig:prompts2}
\end{figure}

\paragraph{Prompts.}
We formulate a prompt tailored for LLaMA-2-13B, enabling it to proficiently paraphrase the given content. The prompt template is shown in Fig.~\ref{fig:prompts}, Fig.~\ref{fig:prompts2}, and Fig.~\ref{fig:prompts3}.

\section{Experimental Results}
\label{app_sec:experimental_results}
Tables \ref{tab:llama2_results} and \ref{tab:llama2_backtranslate_results} show the TPR results of detection when utilizing LLaMA-2-13B as the base model of watermarker. 
Table~\ref{tab:opt_results} shows the TPR results of detection when utilizing OPT-1.3B as the base model of watermarker.
Table~\ref{tab:qwen_paraphrase_results} shows the TPR results of detection when utilizing OPT-1.3B as the base model of watermarker.

Tables \ref{tab:example_weak2weak} and \ref{tab:example_strong2strong} also present several examples of the watermarked texts under different settings.

\begin{figure*}[t]
\newcommand{\imageSize}{0.48\linewidth}
\centering
\begin{subfigure}[t]{\imageSize}
    \centering
    \includegraphics[width=\linewidth]{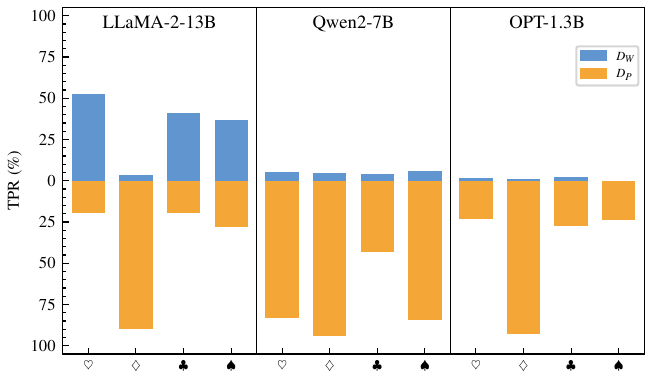}
    \caption{weak $W$, weak $P$}
\end{subfigure}
\begin{subfigure}[t]{\imageSize}
    \centering
    \includegraphics[width=\linewidth]{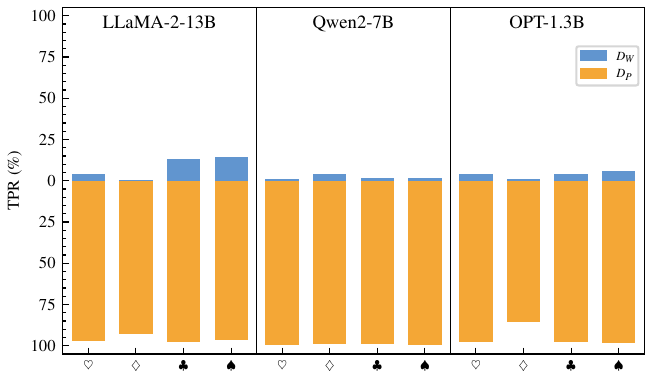}
    \caption{weak $W$, strong $P$}
\end{subfigure}
\begin{subfigure}[t]{\imageSize}
    \centering
    \includegraphics[width=\linewidth]{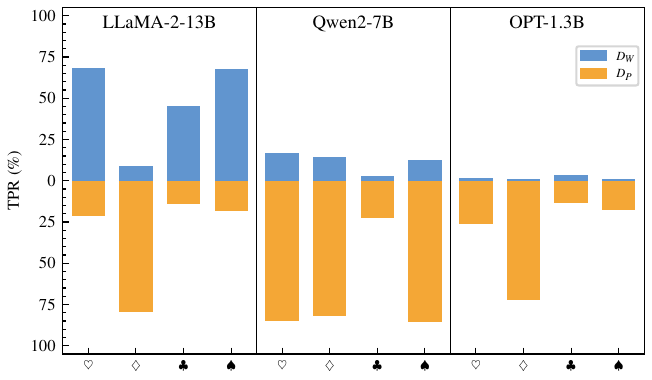}
    \caption{strong $W$, weak $P$}
\end{subfigure}
\begin{subfigure}[t]{\imageSize}
    \centering
    \includegraphics[width=\linewidth]{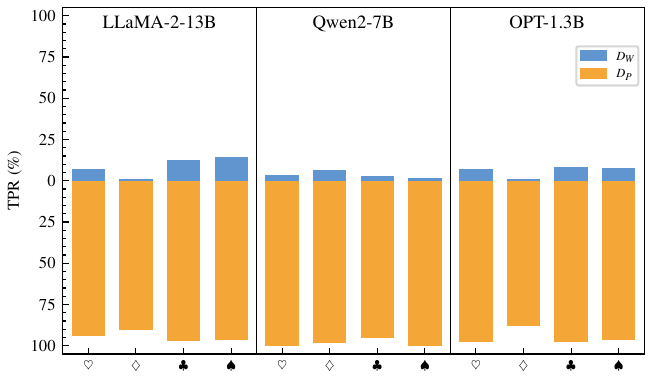}
    \caption{strong $W$, strong $P$}
\end{subfigure}
\caption{
TPR of the \emph{paraphrased} text $T_P$ with different settings and base models as both the watermarker and paraphraser. 
Blue bars represent TPRs detected by the original watermark detector $D_W$, and orange bars are TPRs detected by the collider $D_P$.
Symbols $\heartsuit,\diamondsuit,\clubsuit,\spadesuit$ denote different watermarker-paraphraser pairs as KGW-KGW, KGW-SIR, SIR-PRW and PRW-KGW, respectively.
}
\label{fig:model_compare}
\end{figure*}

\paragraph{Collision can be observed across different base models.}
This observation is supported by the use of LLaMA-2-13B, Qwen2-7B, and OPT-1.3B as the base models, as illustrated in Fig.~\ref{fig:model_compare}. 
The findings suggest that watermark collision is inevitable across different base models, proving its universal applicability as a methodology. 

\paragraph{The semantics of paraphrasing is mostly maintained in weak settings, while strong colliders preserve it to some degree.}
Table~\ref{tab:semantics} shows the similarity of sentence embeddings across various settings, measured by cosine similarity using the \texttt{all-MiniLM-L6-v2} model from the \texttt{sentence-transformer} library.

\section{Scientific Artifacts}
The licenses for all the watermarking methods are listed below:
KGW (Apache 2.0 Licence), SIR (MIT Licence), PRW (MIT Licence). 
The licenses for models are listed below:
LLaMA-2-13B (LLAMA 2 Community License), Qwen2-7B-Instruct (Apache 2.0 Licence), OPT-1.3B (OPT LICENSE).

\begin{table*}[]
\centering
\begin{subtable}[b]{0.48\textwidth}
\centering
\scalebox{0.55}{
\begin{tabular}{c|cc|cc|cc|cc}
\toprule
\textbf{Watermarker} & \multicolumn{4}{c}{\textbf{$\text{KGW}_{weak}$}} & \multicolumn{4}{c}{\textbf{$\text{KGW}_{strong}$}} \\
\midrule
\textbf{Paraphraser} & \multicolumn{2}{c}{\textbf{$\text{KGW}_{weak}$}} & \multicolumn{2}{c}{\textbf{$\text{KGW}_{strong}$}} & \multicolumn{2}{c}{\textbf{$\text{KGW}_{weak}$}} & \multicolumn{2}{c}{\textbf{$\text{KGW}_{strong}$}} \\
\midrule
\textbf{FPR} & \textbf{$D_W$} & \textbf{$D_P$} & \textbf{$D_W$} & \textbf{$D_P$} & \textbf{$D_W$} & \textbf{$D_P$} & \textbf{$D_W$} & \textbf{$D_P$} \\
\midrule
1\% & 52.80 & 19.80 & 4.10 & 97.40 & 68.20 & 21.50 & 7.00 & 94.10 \\
5\% & 66.60 & 34.20 & 14.30 & 99.60 & 72.80 & 32.70 & 16.20 & 97.90 \\
10\% & 71.00 & 41.50 & 22.10 & 99.60 & 75.00 & 41.70 & 26.10 & 99.40 \\
\bottomrule
\end{tabular}
}
\caption{}
\end{subtable}
\hfill
\begin{subtable}[b]{0.48\textwidth}
\centering
\scalebox{0.55}{
\begin{tabular}{c|cc|cc|cc|cc}
\toprule
\textbf{Watermarker} & \multicolumn{4}{c}{\textbf{$\text{KGW}_{weak}$}} & \multicolumn{4}{c}{\textbf{$\text{KGW}_{strong}$}} \\
\midrule
\textbf{Paraphraser} & \multicolumn{2}{c}{\textbf{$\text{SIR}_{weak}$}} & \multicolumn{2}{c}{\textbf{$\text{SIR}_{strong}$}} & \multicolumn{2}{c}{\textbf{$\text{SIR}_{weak}$}} & \multicolumn{2}{c}{\textbf{$\text{SIR}_{strong}$}} \\
\midrule
\textbf{FPR} & \textbf{$D_W$} & \textbf{$D_P$} & \textbf{$D_W$} & \textbf{$D_P$} & \textbf{$D_W$} & \textbf{$D_P$} & \textbf{$D_W$} & \textbf{$D_P$} \\
\midrule
1\% & 3.40 & 90.09 & 0.20 & 92.77 & 9.00 & 79.26 & 0.80 & 90.52 \\
5\% & 5.90 & 92.12 & 2.00 & 94.66 & 14.10 & 89.41 & 2.50 & 93.76 \\
10\% & 7.80 & 93.81 & 3.80 & 95.88 & 17.30 & 92.14 & 4.70 & 95.88 \\
\bottomrule
\end{tabular}
}
\caption{}
\end{subtable}

\begin{subtable}[b]{0.48\textwidth}
\centering
\scalebox{0.55}{
\begin{tabular}{c|cc|cc|cc|cc}
\toprule
\textbf{Watermarker} & \multicolumn{4}{c}{\textbf{$\text{KGW}_{weak}$}} & \multicolumn{4}{c}{\textbf{$\text{KGW}_{strong}$}} \\
\midrule
\textbf{Paraphraser} & \multicolumn{2}{c}{\textbf{$\text{PRW}_{weak}$}} & \multicolumn{2}{c}{\textbf{$\text{PRW}_{strong}$}} & \multicolumn{2}{c}{\textbf{$\text{PRW}_{weak}$}} & \multicolumn{2}{c}{\textbf{$\text{PRW}_{strong}$}} \\
\midrule
\textbf{FPR} & \textbf{$D_W$} & \textbf{$D_P$} & \textbf{$D_W$} & \textbf{$D_P$} & \textbf{$D_W$} & \textbf{$D_P$} & \textbf{$D_W$} & \textbf{$D_P$} \\
\midrule
1\% & 41.10 & 48.00 & 6.20 & 99.90 & 56.00 & 44.40 & 14.80 & 99.70 \\
5\% & 54.40 & 65.70 & 16.70 & 99.90 & 64.70 & 64.30 & 25.40 & 100.00 \\
10\% & 60.20 & 71.70 & 23.70 & 99.90 & 68.50 & 70.80 & 33.00 & 100.00 \\
\bottomrule
\end{tabular}
}
\caption{}
\end{subtable}
\hfill
\begin{subtable}[b]{0.48\textwidth}
\centering
\scalebox{0.55}{
\begin{tabular}{c|cc|cc|cc|cc}
\toprule
\textbf{Watermarker} & \multicolumn{4}{c}{\textbf{$\text{SIR}_{weak}$}} & \multicolumn{4}{c}{\textbf{$\text{SIR}_{strong}$}} \\
\midrule
\textbf{Paraphraser} & \multicolumn{2}{c}{\textbf{$\text{KGW}_{weak}$}} & \multicolumn{2}{c}{\textbf{$\text{KGW}_{strong}$}} & \multicolumn{2}{c}{\textbf{$\text{KGW}_{weak}$}} & \multicolumn{2}{c}{\textbf{$\text{KGW}_{strong}$}} \\
\midrule
\textbf{FPR} & \textbf{$D_W$} & \textbf{$D_P$} & \textbf{$D_W$} & \textbf{$D_P$} & \textbf{$D_W$} & \textbf{$D_P$} & \textbf{$D_W$} & \textbf{$D_P$} \\
\midrule
1\% & 55.74 & 25.10 & 12.45 & 96.70 & 61.55 & 21.00 & 10.98 & 97.40 \\
5\% & 65.97 & 34.80 & 25.86 & 99.20 & 72.95 & 30.90 & 27.56 & 98.80 \\
10\% & 69.73 & 40.30 & 31.22 & 99.50 & 75.44 & 38.70 & 34.23 & 99.40 \\
\bottomrule
\end{tabular}
}
\caption{}
\end{subtable}

\begin{subtable}[b]{0.48\textwidth}
\centering
\scalebox{0.55}{
\begin{tabular}{c|cc|cc|cc|cc}
\toprule
\textbf{Watermarker} & \multicolumn{4}{c}{\textbf{$\text{SIR}_{weak}$}} & \multicolumn{4}{c}{\textbf{$\text{SIR}_{strong}$}} \\
\midrule
\textbf{Paraphraser} & \multicolumn{2}{c}{\textbf{$\text{PRW}_{weak}$}} & \multicolumn{2}{c}{\textbf{$\text{PRW}_{strong}$}} & \multicolumn{2}{c}{\textbf{$\text{PRW}_{weak}$}} & \multicolumn{2}{c}{\textbf{$\text{PRW}_{strong}$}} \\
\midrule
\textbf{FPR} & \textbf{$D_W$} & \textbf{$D_P$} & \textbf{$D_W$} & \textbf{$D_P$} & \textbf{$D_W$} & \textbf{$D_P$} & \textbf{$D_W$} & \textbf{$D_P$} \\
\midrule
1\% & 41.05 & 19.80 & 13.22 & 97.50 & 45.58 & 14.00 & 12.58 & 97.10 \\
5\% & 52.67 & 56.60 & 22.43 & 100.00 & 59.98 & 45.90 & 25.48 & 99.70 \\
10\% & 56.17 & 67.30 & 27.63 & 100.00 & 63.17 & 59.00 & 30.49 & 99.90 \\
\bottomrule
\end{tabular}
}
\caption{}
\end{subtable}
\hfill
\begin{subtable}[b]{0.48\textwidth}
\centering
\scalebox{0.55}{
\begin{tabular}{c|cc|cc|cc|cc}
\toprule
\textbf{Watermarker} & \multicolumn{4}{c}{\textbf{$\text{PRW}_{weak}$}} & \multicolumn{4}{c}{\textbf{$\text{PRW}_{strong}$}} \\
\midrule
\textbf{Paraphraser} & \multicolumn{2}{c}{\textbf{$\text{KGW}_{weak}$}} & \multicolumn{2}{c}{\textbf{$\text{KGW}_{strong}$}} & \multicolumn{2}{c}{\textbf{$\text{KGW}_{weak}$}} & \multicolumn{2}{c}{\textbf{$\text{KGW}_{strong}$}} \\
\midrule
\textbf{FPR} & \textbf{$D_W$} & \textbf{$D_P$} & \textbf{$D_W$} & \textbf{$D_P$} & \textbf{$D_W$} & \textbf{$D_P$} & \textbf{$D_W$} & \textbf{$D_P$} \\
\midrule
1\% & 37.00 & 28.20 & 14.60 & 96.70 & 67.90 & 18.30 & 14.60 & 96.70 \\
5\% & 56.40 & 38.60 & 22.30 & 99.20 & 73.50 & 29.70 & 22.30 & 99.20 \\
10\% & 65.80 & 45.70 & 30.20 & 99.40 & 76.50 & 38.70 & 30.20 & 99.40 \\
\bottomrule
\end{tabular}
}
\caption{}
\end{subtable}

\begin{subtable}[b]{0.48\textwidth}
\centering
\scalebox{0.55}{
\begin{tabular}{c|cc|cc|cc|cc}
\toprule
\textbf{Watermarker} & \multicolumn{4}{c}{\textbf{$\text{PRW}_{weak}$}} & \multicolumn{4}{c}{\textbf{$\text{PRW}_{strong}$}} \\
\midrule
\textbf{Paraphraser} & \multicolumn{2}{c}{\textbf{$\text{SIR}_{weak}$}} & \multicolumn{2}{c}{\textbf{$\text{SIR}_{strong}$}} & \multicolumn{2}{c}{\textbf{$\text{SIR}_{weak}$}} & \multicolumn{2}{c}{\textbf{$\text{SIR}_{strong}$}} \\
\midrule
\textbf{FPR} & \textbf{$D_W$} & \textbf{$D_P$} & \textbf{$D_W$} & \textbf{$D_P$} & \textbf{$D_W$} & \textbf{$D_P$} & \textbf{$D_W$} & \textbf{$D_P$} \\
\midrule
1\% & 22.30 & 79.56 & 29.40 & 91.91 & 32.00 & 65.31 & 37.10 & 79.00 \\
5\% & 31.80 & 91.00 & 37.70 & 94.43 & 42.20 & 87.78 & 47.10 & 90.89 \\
10\% & 39.20 & 92.56 & 42.10 & 95.63 & 47.90 & 91.19 & 51.20 & 93.22 \\
\bottomrule
\end{tabular}
}
\caption{}
\end{subtable}
\hfill
\begin{subtable}[b]{0.48\textwidth}
\centering
\scalebox{0.55}{
\begin{tabular}{c|cc|cc|cc|cc}
\toprule
\textbf{Watermarker} & \multicolumn{4}{c}{\textbf{$\text{PRW}_{weak}$}} & \multicolumn{4}{c}{\textbf{$\text{PRW}_{strong}$}} \\
\midrule
\textbf{Paraphraser} & \multicolumn{2}{c}{\textbf{$\text{PRW}_{weak}$}} & \multicolumn{2}{c}{\textbf{$\text{PRW}_{strong}$}} & \multicolumn{2}{c}{\textbf{$\text{PRW}_{weak}$}} & \multicolumn{2}{c}{\textbf{$\text{PRW}_{strong}$}} \\
\midrule
\textbf{FPR} & \textbf{$D_W$} & \textbf{$D_P$} & \textbf{$D_W$} & \textbf{$D_P$} & \textbf{$D_W$} & \textbf{$D_P$} & \textbf{$D_W$} & \textbf{$D_P$} \\
\midrule
1\% & 26.60 & 41.20 & 9.30 & 99.90 & 44.20 & 41.70 & 11.20 & 99.30 \\
5\% & 37.80 & 67.20 & 14.30 & 100.00 & 50.60 & 58.50 & 18.60 & 99.90 \\
10\% & 45.30 & 72.00 & 18.60 & 100.00 & 55.50 & 65.20 & 24.90 & 99.90 \\
\bottomrule
\end{tabular}
}
\caption{}
\end{subtable}

\caption{TPR of the paraphrased text $T_P$ with dual watermarks when utilizing LLaMA-2-13B as the base model of watermarker. FPR is set to $1\%$, $2\%$ \& $5\%$, respectively.}
\label{tab:llama2_results}
\end{table*}

\begin{table*}

\begin{subtable}[b]{0.48\textwidth}
\centering
\scalebox{0.55}{
\begin{tabular}{c|cc|cc|cc|cc}
\toprule
\textbf{Watermarker} & \multicolumn{4}{c}{\textbf{$\text{KGW}_{weak}$}} & \multicolumn{4}{c}{\textbf{$\text{KGW}_{strong}$}} \\
\midrule
\textbf{Paraphraser} & \multicolumn{2}{c}{\textbf{$\text{KGW}_{weak}$}} & \multicolumn{2}{c}{\textbf{$\text{KGW}_{strong}$}} & \multicolumn{2}{c}{\textbf{$\text{KGW}_{weak}$}} & \multicolumn{2}{c}{\textbf{$\text{KGW}_{strong}$}} \\
\midrule
\textbf{FPR} & \textbf{$D_W$} & \textbf{$D_P$} & \textbf{$D_W$} & \textbf{$D_P$} & \textbf{$D_W$} & \textbf{$D_P$} & \textbf{$D_W$} & \textbf{$D_P$} \\
\midrule
1\% & 1.40 & 23.40 & 4.30 & 98.00 & 1.50 & 26.30 & 6.90 & 97.90 \\
5\% & 10.90 & 34.30 & 17.60 & 99.40 & 7.30 & 36.50 & 19.90 & 99.20 \\
10\% & 17.20 & 44.10 & 25.60 & 99.60 & 15.30 & 44.20 & 28.30 & 99.40 \\
\bottomrule
\end{tabular}
}
\caption{}
\end{subtable}
\hfill
\begin{subtable}[b]{0.48\textwidth}
\centering
\scalebox{0.55}{
\begin{tabular}{c|cc|cc|cc|cc}
\toprule
\textbf{Watermarker} & \multicolumn{4}{c}{\textbf{$\text{KGW}_{weak}$}} & \multicolumn{4}{c}{\textbf{$\text{KGW}_{strong}$}} \\
\midrule
\textbf{Paraphraser} & \multicolumn{2}{c}{\textbf{$\text{SIR}_{weak}$}} & \multicolumn{2}{c}{\textbf{$\text{SIR}_{strong}$}} & \multicolumn{2}{c}{\textbf{$\text{SIR}_{weak}$}} & \multicolumn{2}{c}{\textbf{$\text{SIR}_{strong}$}} \\
\midrule
\textbf{FPR} & \textbf{$D_W$} & \textbf{$D_P$} & \textbf{$D_W$} & \textbf{$D_P$} & \textbf{$D_W$} & \textbf{$D_P$} & \textbf{$D_W$} & \textbf{$D_P$} \\
\midrule
1\% & 0.80 & 92.81 & 0.90 & 85.32 & 0.90 & 72.27 & 1.20 & 87.80 \\
5\% & 3.60 & 94.72 & 3.80 & 91.24 & 3.80 & 87.62 & 4.10 & 93.63 \\
10\% & 6.70 & 96.18 & 6.70 & 93.65 & 7.40 & 91.60 & 8.10 & 95.49 \\
\bottomrule
\end{tabular}
}
\caption{}
\end{subtable}

\begin{subtable}[b]{0.48\textwidth}
\centering
\scalebox{0.55}{
\begin{tabular}{c|cc|cc|cc|cc}
\toprule
\textbf{Watermarker} & \multicolumn{4}{c}{\textbf{$\text{KGW}_{weak}$}} & \multicolumn{4}{c}{\textbf{$\text{KGW}_{strong}$}} \\
\midrule
\textbf{Paraphraser} & \multicolumn{2}{c}{\textbf{$\text{PRW}_{weak}$}} & \multicolumn{2}{c}{\textbf{$\text{PRW}_{strong}$}} & \multicolumn{2}{c}{\textbf{$\text{PRW}_{weak}$}} & \multicolumn{2}{c}{\textbf{$\text{PRW}_{strong}$}} \\
\midrule
\textbf{FPR} & \textbf{$D_W$} & \textbf{$D_P$} & \textbf{$D_W$} & \textbf{$D_P$} & \textbf{$D_W$} & \textbf{$D_P$} & \textbf{$D_W$} & \textbf{$D_P$} \\
\midrule
1\% & 5.00 & 36.80 & 6.90 & 99.80 & 5.80 & 14.70 & 8.80 & 97.40 \\
5\% & 21.10 & 56.10 & 22.30 & 100.00 & 17.80 & 52.30 & 22.00 & 99.80 \\
10\% & 34.30 & 66.20 & 31.60 & 100.00 & 29.00 & 63.00 & 31.50 & 100.00 \\
\bottomrule
\end{tabular}
}
\caption{}
\end{subtable}
\hfill
\begin{subtable}[b]{0.48\textwidth}
\centering
\scalebox{0.55}{
\begin{tabular}{c|cc|cc|cc|cc}
\toprule
\textbf{Watermarker} & \multicolumn{4}{c}{\textbf{$\text{SIR}_{weak}$}} & \multicolumn{4}{c}{\textbf{$\text{SIR}_{strong}$}} \\
\midrule
\textbf{Paraphraser} & \multicolumn{2}{c}{\textbf{$\text{KGW}_{weak}$}} & \multicolumn{2}{c}{\textbf{$\text{KGW}_{strong}$}} & \multicolumn{2}{c}{\textbf{$\text{KGW}_{weak}$}} & \multicolumn{2}{c}{\textbf{$\text{KGW}_{strong}$}} \\
\midrule
\textbf{FPR} & \textbf{$D_W$} & \textbf{$D_P$} & \textbf{$D_W$} & \textbf{$D_P$} & \textbf{$D_W$} & \textbf{$D_P$} & \textbf{$D_W$} & \textbf{$D_P$} \\
\midrule
1\% & 1.25 & 19.30 & 5.17 & 96.40 & 2.61 & 17.00 & 6.52 & 93.70 \\
5\% & 4.05 & 32.30 & 15.82 & 98.30 & 6.58 & 26.90 & 16.04 & 98.80 \\
10\% & 6.76 & 40.50 & 20.34 & 98.70 & 10.44 & 35.20 & 20.32 & 99.30 \\
\bottomrule
\end{tabular}
}
\caption{}
\end{subtable}

\begin{subtable}[b]{0.48\textwidth}
\centering
\scalebox{0.55}{
\begin{tabular}{c|cc|cc|cc|cc}
\toprule
\textbf{Watermarker} & \multicolumn{4}{c}{\textbf{$\text{SIR}_{weak}$}} & \multicolumn{4}{c}{\textbf{$\text{SIR}_{strong}$}} \\
\midrule
\textbf{Paraphraser} & \multicolumn{2}{c}{\textbf{$\text{PRW}_{weak}$}} & \multicolumn{2}{c}{\textbf{$\text{PRW}_{strong}$}} & \multicolumn{2}{c}{\textbf{$\text{PRW}_{weak}$}} & \multicolumn{2}{c}{\textbf{$\text{PRW}_{strong}$}} \\
\midrule
\textbf{FPR} & \textbf{$D_W$} & \textbf{$D_P$} & \textbf{$D_W$} & \textbf{$D_P$} & \textbf{$D_W$} & \textbf{$D_P$} & \textbf{$D_W$} & \textbf{$D_P$} \\
\midrule
1\% & 2.52 & 27.20 & 4.26 & 97.70 & 3.76 & 13.70 & 8.25 & 97.80 \\
5\% & 5.45 & 52.20 & 10.93 & 99.70 & 8.57 & 46.00 & 13.14 & 100.00 \\
10\% & 8.28 & 59.70 & 13.77 & 99.80 & 11.60 & 55.60 & 15.53 & 100.00 \\
\bottomrule
\end{tabular}
}
\caption{}
\end{subtable}
\hfill
\begin{subtable}[b]{0.48\textwidth}
\centering
\scalebox{0.55}{
\begin{tabular}{c|cc|cc|cc|cc}
\toprule
\textbf{Watermarker} & \multicolumn{4}{c}{\textbf{$\text{PRW}_{weak}$}} & \multicolumn{4}{c}{\textbf{$\text{PRW}_{strong}$}} \\
\midrule
\textbf{Paraphraser} & \multicolumn{2}{c}{\textbf{$\text{KGW}_{weak}$}} & \multicolumn{2}{c}{\textbf{$\text{KGW}_{strong}$}} & \multicolumn{2}{c}{\textbf{$\text{KGW}_{weak}$}} & \multicolumn{2}{c}{\textbf{$\text{KGW}_{strong}$}} \\
\midrule
\textbf{FPR} & \textbf{$D_W$} & \textbf{$D_P$} & \textbf{$D_W$} & \textbf{$D_P$} & \textbf{$D_W$} & \textbf{$D_P$} & \textbf{$D_W$} & \textbf{$D_P$} \\
\midrule
1\% & 0.10 & 23.50 & 5.70 & 98.60 & 1.30 & 17.50 & 7.80 & 96.80 \\
5\% & 2.90 & 35.60 & 17.30 & 99.40 & 6.30 & 29.70 & 21.80 & 98.60 \\
10\% & 13.60 & 44.20 & 34.00 & 99.50 & 11.70 & 42.10 & 29.60 & 98.70 \\
\bottomrule
\end{tabular}
}
\caption{}
\end{subtable}

\begin{subtable}[b]{0.48\textwidth}
\centering
\scalebox{0.55}{
\begin{tabular}{c|cc|cc|cc|cc}
\toprule
\textbf{Watermarker} & \multicolumn{4}{c}{\textbf{$\text{PRW}_{weak}$}} & \multicolumn{4}{c}{\textbf{$\text{PRW}_{strong}$}} \\
\midrule
\textbf{Paraphraser} & \multicolumn{2}{c}{\textbf{$\text{SIR}_{weak}$}} & \multicolumn{2}{c}{\textbf{$\text{SIR}_{strong}$}} & \multicolumn{2}{c}{\textbf{$\text{SIR}_{weak}$}} & \multicolumn{2}{c}{\textbf{$\text{SIR}_{strong}$}} \\
\midrule
\textbf{FPR} & \textbf{$D_W$} & \textbf{$D_P$} & \textbf{$D_W$} & \textbf{$D_P$} & \textbf{$D_W$} & \textbf{$D_P$} & \textbf{$D_W$} & \textbf{$D_P$} \\
\midrule
1\% & 10.60 & 87.23 & 24.90 & 88.96 & 14.00 & 78.84 & 24.30 & 88.62 \\
5\% & 24.30 & 91.38 & 38.60 & 93.93 & 30.30 & 89.81 & 38.50 & 93.53 \\
10\% & 38.90 & 92.50 & 48.60 & 94.92 & 36.60 & 92.39 & 43.60 & 94.53 \\
\bottomrule
\end{tabular}
}
\caption{}
\end{subtable}
\hfill
\begin{subtable}[b]{0.48\textwidth}
\centering
\scalebox{0.55}{
\begin{tabular}{c|cc|cc|cc|cc}
\toprule
\textbf{Watermarker} & \multicolumn{4}{c}{\textbf{$\text{PRW}_{weak}$}} & \multicolumn{4}{c}{\textbf{$\text{PRW}_{strong}$}} \\
\midrule
\textbf{Paraphraser} & \multicolumn{2}{c}{\textbf{$\text{PRW}_{weak}$}} & \multicolumn{2}{c}{\textbf{$\text{PRW}_{strong}$}} & \multicolumn{2}{c}{\textbf{$\text{PRW}_{weak}$}} & \multicolumn{2}{c}{\textbf{$\text{PRW}_{strong}$}} \\
\midrule
\textbf{FPR} & \textbf{$D_W$} & \textbf{$D_P$} & \textbf{$D_W$} & \textbf{$D_P$} & \textbf{$D_W$} & \textbf{$D_P$} & \textbf{$D_W$} & \textbf{$D_P$} \\
\midrule
1\% & 0.80 & 36.10 & 4.70 & 99.60 & 1.00 & 11.60 & 6.10 & 93.30 \\
5\% & 3.70 & 57.80 & 10.70 & 100.00 & 6.50 & 52.10 & 13.80 & 100.00 \\
10\% & 10.30 & 66.00 & 19.90 & 100.00 & 11.60 & 63.20 & 18.40 & 100.00 \\
\bottomrule
\end{tabular}
}
\caption{}
\end{subtable}

\caption{TPR of the paraphrased text $T_P$ with dual watermarks when utilizing OPT-1.3B as the base model of watermarker. FPR is set to $1\%$, $2\%$ \& $5\%$, respectively.}
\label{tab:opt_results}
\end{table*}

\begin{table*}[]
\centering
\begin{subtable}[b]{0.48\textwidth}
\centering
\begin{tabular}{c|c|c|c|c}
\toprule
\diagbox{$W$}{$P$} & $P'$ & \weak{KGW} & \weak{PRW} & \weak{SIR} \\
\midrule
\weak{KGW} & 0.879 & 0.852 & 0.858 & 0.790 \\
\weak{PRW} & 0.888 & 0.864 & 0.872 & 0.781 \\
\weak{SIR} & 0.867 & 0.852 & 0.873 & - \\
\bottomrule
\end{tabular}
\caption{}
\end{subtable}
\hfill
\begin{subtable}[b]{0.48\textwidth}
\centering
\begin{tabular}{c|c|c|c|c}
\toprule
\diagbox{$W$}{$P$} & $P'$ & \strong{KGW} & \strong{PRW} & \strong{SIR} \\
\midrule
\weak{KGW} & 0.879 & 0.714 & 0.722 & 0.699 \\
\weak{PRW} & 0.888 & 0.721 & 0.726 & 0.709 \\
\weak{SIR} & 0.867 & 0.730 & 0.748 & - \\
\bottomrule
\end{tabular}
\caption{}
\end{subtable}

\begin{subtable}[b]{0.48\textwidth}
\centering
\begin{tabular}{c|c|c|c|c}
\toprule
\diagbox{$W$}{$P$} & $P'$ & \weak{KGW} & \weak{PRW} & \weak{SIR} \\
\midrule
\strong{KGW} & 0.874 & 0.847 & 0.857 & 0.788 \\
\strong{PRW} & 0.887 & 0.849 & 0.867 & 0.776 \\
\strong{SIR} & 0.883 & 0.851 & 0.868 & - \\
\bottomrule
\end{tabular}
\caption{}
\end{subtable}
\hfill
\begin{subtable}[b]{0.48\textwidth}
\centering
\begin{tabular}{c|c|c|c|c}
\toprule
\diagbox{$W$}{$P$} & $P'$ & \strong{KGW} & \strong{PRW} & \strong{SIR} \\
\midrule
\strong{KGW} & 0.874 & 0.706 & 0.725 & 0.696 \\
\strong{PRW} & 0.887 & 0.719 & 0.721 & 0.689 \\
\strong{SIR} & 0.883 & 0.721 & 0.737 & - \\
\bottomrule
\end{tabular}
\caption{}
\end{subtable}

\caption{Semantic similarity between paraphrased text and original text of LLaMA-2-13B.}
\label{tab:semantics}
\end{table*}


\begin{sidewaystable*}[]
\newcommand{\scaleSize}{0.60}

\begin{subtable}[]{0.99\textwidth}
\scalebox{\scaleSize}{
\begin{tabular}{c|cc|cc|cc|cc|cc|cc|cc}
\toprule
\textbf{Watermarker} & \multicolumn{14}{c}{\textbf{\weak{KGW}}} \\
\midrule
\textbf{Translator} & \multicolumn{2}{c}{No Watermark} & \multicolumn{2}{c}{\textbf{\weak{KGW}}} & \multicolumn{2}{c}{\textbf{\strong{KGW}}} & \multicolumn{2}{c}{\textbf{\weak{PRW}}} & \multicolumn{2}{c}{\textbf{\strong{PRW}}} & \multicolumn{2}{c}{\textbf{\weak{SIR}}} & \multicolumn{2}{c}{\textbf{\strong{SIR}}} \\
\textbf{FPR} & \textbf{Gen TPR} & \multicolumn{1}{c}{\textbf{$D_W$}} & \multicolumn{1}{c}{\textbf{Old TPR}} & \multicolumn{1}{c}{\textbf{New TPR}} & \multicolumn{1}{c}{\textbf{Old TPR}} & \multicolumn{1}{c}{\textbf{New TPR}} & \multicolumn{1}{c}{\textbf{Old TPR}} & \multicolumn{1}{c}{\textbf{New TPR}} & \multicolumn{1}{c}{\textbf{Old TPR}} & \multicolumn{1}{c}{\textbf{New TPR}} & \multicolumn{1}{c}{\textbf{Old TPR}} & \multicolumn{1}{c}{\textbf{New TPR}} & \multicolumn{1}{c}{\textbf{Old TPR}} & \multicolumn{1}{c}{\textbf{New TPR}} \\
\midrule
1\% & \multicolumn{1}{r}{99.9} & 44.8 & 41.9 & 9.2 & 28.5 & 69.5 & 34.2 & 13.7 & 18.5 & 58.5 & 31.4 & 5.15 & 3.9 & 93.5 \\
5\% & \multicolumn{1}{r}{100} & 69.7 & 66.5 & 19.6 & 51.2 & 79.1 & 59.8 & 35.8 & 40.7 & 82.8 & 56.8 & 30.37 & 14.6 & 95.56 \\
10\% & \multicolumn{1}{r}{100} & 78.4 & 75.2 & 30.3 & 63 & 85.5 & 70.3 & 50.1 & 55 & 89.9 & 68.4 & 46.92 & 21.8 & 96.08 \\
\bottomrule
\end{tabular}}
\caption{}
\end{subtable}

\hfill
\hfill

\begin{subtable}[]{0.99\textwidth}
\scalebox{\scaleSize}{
\begin{tabular}{c|cc|cc|cc|cc|cc|cc|cc}
\toprule
\textbf{Watermarker} & \multicolumn{14}{c}{\textbf{\strong{KGW}}} \\
\midrule
\textbf{Translator} & \multicolumn{2}{c}{No Watermark} & \multicolumn{2}{c}{\textbf{\weak{KGW}}} & \multicolumn{2}{c}{\textbf{\strong{KGW}}} & \multicolumn{2}{c}{\textbf{\weak{PRW}}} & \multicolumn{2}{c}{\textbf{\strong{PRW}}} & \multicolumn{2}{c}{\textbf{\weak{SIR}}} & \multicolumn{2}{c}{\textbf{\strong{SIR}}} \\
\textbf{FPR} & \textbf{Gen TPR} & \multicolumn{1}{c}{\textbf{$D_W$}} & \multicolumn{1}{c}{\textbf{Old TPR}} & \multicolumn{1}{c}{\textbf{New TPR}} & \multicolumn{1}{c}{\textbf{Old TPR}} & \multicolumn{1}{c}{\textbf{New TPR}} & \multicolumn{1}{c}{\textbf{Old TPR}} & \multicolumn{1}{c}{\textbf{New TPR}} & \multicolumn{1}{c}{\textbf{Old TPR}} & \multicolumn{1}{c}{\textbf{New TPR}} & \multicolumn{1}{c}{\textbf{Old TPR}} & \multicolumn{1}{c}{\textbf{New TPR}} & \multicolumn{1}{c}{\textbf{Old TPR}} & \multicolumn{1}{c}{\textbf{New TPR}} \\
\midrule
1\% & \multicolumn{1}{r}{100} & 69.3 & 67.8 & 8.1 & 50 & 68.9 & 61.9 & 13.6 & 38.8 & 53 & 55.3 & 4.09 & 7.5 & 91.7 \\
5\% & \multicolumn{1}{r}{100} & 80.7 & 80.1 & 25.3 & 66.8 & 83 & 76 & 40.6 & 56.7 & 84.3 & 70.8 & 36.09 & 20.8 & 95.95 \\
10\% & \multicolumn{1}{r}{100} & 86.4 & 84.1 & 35.2 & 74.5 & 88.2 & 81.8 & 58.4 & 65.9 & 92.5 & 76.9 & 52.25 & 29.6 & 96.89 \\
\bottomrule
\end{tabular}}
\caption{}
\end{subtable}

\hfill
\hfill

\begin{subtable}[]{0.99\textwidth}
\scalebox{\scaleSize}{
\begin{tabular}{c|cc|cc|cc|cc|cc|cc|cc}
\toprule
\textbf{Watermarker} & \multicolumn{14}{c}{\textbf{\weak{PRW}}} \\
\midrule
\textbf{Translator} & \multicolumn{2}{c}{No Watermark} & \multicolumn{2}{c}{\textbf{\weak{KGW}}} & \multicolumn{2}{c}{\textbf{\strong{KGW}}} & \multicolumn{2}{c}{\textbf{\weak{PRW}}} & \multicolumn{2}{c}{\textbf{\strong{PRW}}} & \multicolumn{2}{c}{\textbf{\weak{SIR}}} & \multicolumn{2}{c}{\textbf{\strong{SIR}}} \\
\textbf{FPR} & \textbf{Gen TPR} & \multicolumn{1}{c}{\textbf{$D_W$}} & \multicolumn{1}{c}{\textbf{Old TPR}} & \multicolumn{1}{c}{\textbf{New TPR}} & \multicolumn{1}{c}{\textbf{Old TPR}} & \multicolumn{1}{c}{\textbf{New TPR}} & \multicolumn{1}{c}{\textbf{Old TPR}} & \multicolumn{1}{c}{\textbf{New TPR}} & \multicolumn{1}{c}{\textbf{Old TPR}} & \multicolumn{1}{c}{\textbf{New TPR}} & \multicolumn{1}{c}{\textbf{Old TPR}} & \multicolumn{1}{c}{\textbf{New TPR}} & \multicolumn{1}{c}{\textbf{Old TPR}} & \multicolumn{1}{c}{\textbf{New TPR}} \\
\midrule
1\% & \multicolumn{1}{r}{95.4} & 32.9 & \textbf{34.1} & 7.5 & 23 & 68 & 21.8 & 12.7 & 14.3 & 54.6 & 25.2 & 5.73 & 10.8 & 92.87 \\
5\% & \multicolumn{1}{r}{100} & 59.3 & \textbf{61.1} & 20.6 & 46 & 81 & 45.4 & 36 & 31.1 & 78.9 & 49.1 & 25.86 & 18.7 & 95.56 \\
10\% & \multicolumn{1}{r}{100} & 78 & 77.7 & 32 & 65.1 & 86.3 & 67.3 & 51.7 & 50.5 & 88.4 & 69.1 & 44.67 & 31.3 & 96.49 \\
\bottomrule
\end{tabular}}
\caption{}
\end{subtable}

\hfill
\hfill

\begin{subtable}[]{0.99\textwidth}
\scalebox{\scaleSize}{
\begin{tabular}{c|cc|cc|cc|cc|cc|cc|cc}
\toprule
\textbf{Watermarker} & \multicolumn{14}{c}{\textbf{\strong{PRW}}} \\
\midrule
\textbf{Translator} & \multicolumn{2}{c}{No Watermark} & \multicolumn{2}{c}{\textbf{\weak{KGW}}} & \multicolumn{2}{c}{\textbf{\strong{KGW}}} & \multicolumn{2}{c}{\textbf{\weak{PRW}}} & \multicolumn{2}{c}{\textbf{\strong{PRW}}} & \multicolumn{2}{c}{\textbf{\weak{SIR}}} & \multicolumn{2}{c}{\textbf{\strong{SIR}}} \\
\midrule
\textbf{FPR} & \textbf{Gen TPR} & \multicolumn{1}{c}{\textbf{$D_W$}} & \multicolumn{1}{c}{\textbf{Old TPR}} & \multicolumn{1}{c}{\textbf{New TPR}} & \multicolumn{1}{c}{\textbf{Old TPR}} & \multicolumn{1}{c}{\textbf{New TPR}} & \multicolumn{1}{c}{\textbf{Old TPR}} & \multicolumn{1}{c}{\textbf{New TPR}} & \multicolumn{1}{c}{\textbf{Old TPR}} & \multicolumn{1}{c}{\textbf{New TPR}} & \multicolumn{1}{c}{\textbf{Old TPR}} & \multicolumn{1}{c}{\textbf{New TPR}} & \multicolumn{1}{c}{\textbf{Old TPR}} & \multicolumn{1}{c}{\textbf{New TPR}} \\
\midrule
1\% & \multicolumn{1}{r}{99.7} & 63.9 & \textbf{64.8} & 9.6 & 50.1 & 67.9 & 55.2 & 10.2 & 36.6 & 42.4 & 50.6 & 3.56 & 20.1 & 90.44 \\
5\% & \multicolumn{1}{r}{100} & 80.2 & \textbf{81.1} & 22.5 & 71.2 & 81 & 76.9 & 32.2 & 57.7 & 75 & 72.8 & 27.7 & 33.6 & 94.49 \\
10\% & \multicolumn{1}{r}{100} & 87 & 86.8 & 33.7 & 80.5 & 87 & 83.4 & 52.8 & 68.3 & 89 & 80.7 & 46.95 & 45.1 & 95.43 \\
\bottomrule
\end{tabular}}
\caption{}
\end{subtable}

\hfill

\begin{subtable}[]{0.99\textwidth}
\scalebox{\scaleSize}{
\begin{tabular}{c|cc|cc|cc|cc|cc|cc|cc}
\toprule
\textbf{Watermarker} & \multicolumn{14}{c}{\textbf{\weak{SIR}}} \\
\midrule
\textbf{Translator} & \multicolumn{2}{c}{No Watermark} & \multicolumn{2}{c}{\textbf{\weak{KGW}}} & \multicolumn{2}{c}{\textbf{\strong{KGW}}} & \multicolumn{2}{c}{\textbf{\weak{PRW}}} & \multicolumn{2}{c}{\textbf{\strong{PRW}}} & \multicolumn{2}{c}{\textbf{\weak{SIR}}} & \multicolumn{2}{c}{\textbf{\strong{SIR}}} \\
\midrule
\textbf{FPR} & \textbf{Gen TPR} & \multicolumn{1}{c}{\textbf{$D_W$}} & \multicolumn{1}{c}{\textbf{Old TPR}} & \multicolumn{1}{c}{\textbf{New TPR}} & \multicolumn{1}{c}{\textbf{Old TPR}} & \multicolumn{1}{c}{\textbf{New TPR}} & \multicolumn{1}{c}{\textbf{Old TPR}} & \multicolumn{1}{c}{\textbf{New TPR}} & \multicolumn{1}{c}{\textbf{Old TPR}} & \multicolumn{1}{c}{\textbf{New TPR}} & \multicolumn{1}{c}{\textbf{Old TPR}} & \multicolumn{1}{c}{\textbf{New TPR}} & \multicolumn{1}{c}{\textbf{Old TPR}} & \multicolumn{1}{c}{\textbf{New TPR}} \\
\midrule
1\% & \multicolumn{1}{r}{87.9} & 5.6 & \textbf{4.75} & 7.7 & 4.28 & 67.4 & \textbf{6.12} & 11.5 & 4.4 & 41.6 & - & - & - & - \\
5\% & \multicolumn{1}{r}{95.3} & 18.1 & \textbf{14.26} & 24.7 & 16.9 & 84 & \textbf{22.67} & 31 & 17.3 & 72.6 & - & - & - & - \\
10\% & \multicolumn{1}{r}{96.6} & 26.3 & 21.74 & 37.7 & 24.75 & 89.8 & \textbf{32.4} & 52.8 & 25.7 & 86.9 & - & - & - & - \\
\bottomrule
\end{tabular}}
\caption{}
\end{subtable}

\hfill

\begin{subtable}[]{0.99\textwidth}
\scalebox{\scaleSize}{
\begin{tabular}{c|cc|cc|cc|cc|cc|cc|cc}
\toprule
\textbf{Watermarker} & \multicolumn{14}{c}{\textbf{\strong{SIR}}} \\
\midrule
\textbf{Translator} & \multicolumn{2}{c}{No Watermark} & \multicolumn{2}{c}{\textbf{\weak{KGW}}} & \multicolumn{2}{c}{\textbf{\strong{KGW}}} & \multicolumn{2}{c}{\textbf{\weak{PRW}}} & \multicolumn{2}{c}{\textbf{\strong{PRW}}} & \multicolumn{2}{c}{\textbf{\weak{SIR}}} & \multicolumn{2}{c}{\textbf{\strong{SIR}}} \\
\midrule
\textbf{FPR} & \textbf{Gen TPR} & \multicolumn{1}{c}{\textbf{$D_W$}} & \multicolumn{1}{c}{\textbf{Old TPR}} & \multicolumn{1}{c}{\textbf{New TPR}} & \multicolumn{1}{c}{\textbf{Old TPR}} & \multicolumn{1}{c}{\textbf{New TPR}} & \multicolumn{1}{c}{\textbf{Old TPR}} & \multicolumn{1}{c}{\textbf{New TPR}} & \multicolumn{1}{c}{\textbf{Old TPR}} & \multicolumn{1}{c}{\textbf{New TPR}} & \multicolumn{1}{c}{\textbf{Old TPR}} & \multicolumn{1}{c}{\textbf{New TPR}} & \multicolumn{1}{c}{\textbf{Old TPR}} & \multicolumn{1}{c}{\textbf{New TPR}} \\
\midrule
1\% & \multicolumn{1}{r}{94.3} & 3 & \textbf{1.74} & 9.4 & \textbf{4.15} & 74.4 & \textbf{3.82} & 6.7 & 2.51 & 22.5 & - & - & - & - \\
5\% & \multicolumn{1}{r}{97.8} & 16.2 & \textbf{14.23} & 30.4 & \textbf{17.2} & 86.9 & \textbf{19.98} & 34.4 & 15.15 & 73 & - & - & - & - \\
10\% & \multicolumn{1}{r}{98.3} & 24.5 & 23.75 & 42.2 & \textbf{27.98} & 91.2 & \textbf{28.71} & 51.5 & 23.87 & 84.5 & - & - & - & - \\
\bottomrule
\end{tabular}}
\caption{}
\end{subtable}

\caption{TPR of the back-translation text $T_P$ with dual watermarks when utilizing LLaMA-2-13B as the base model of watermarker. FPR is set to $1\%$, $2\%$ \& $5\%$, respectively.}

\label{tab:llama2_backtranslate_results}
\end{sidewaystable*}

\begin{sidewaystable*}[]
\newcommand{\scaleSize}{0.66}

\begin{subtable}[]{0.99\textwidth}
\scalebox{\scaleSize}{
\begin{tabular}{c|cc|cc|cc|cc|cc|cc|cc}
\toprule
\textbf{Watermarker} & \multicolumn{14}{c}{\textbf{\weak{KGW}}} \\
\midrule
\textbf{Translator} & \multicolumn{2}{c}{No Watermark} & \multicolumn{2}{c}{\textbf{\weak{KGW}}} & \multicolumn{2}{c}{\textbf{\strong{KGW}}} & \multicolumn{2}{c}{\textbf{\weak{PRW}}} & \multicolumn{2}{c}{\textbf{\strong{PRW}}} & \multicolumn{2}{c}{\textbf{\weak{SIR}}} & \multicolumn{2}{c}{\textbf{\strong{SIR}}} \\
\textbf{FPR} & \textbf{Gen TPR} & \multicolumn{1}{c}{\textbf{$D_W$}} & \multicolumn{1}{c}{\textbf{Old TPR}} & \multicolumn{1}{c}{\textbf{New TPR}} & \multicolumn{1}{c}{\textbf{Old TPR}} & \multicolumn{1}{c}{\textbf{New TPR}} & \multicolumn{1}{c}{\textbf{Old TPR}} & \multicolumn{1}{c}{\textbf{New TPR}} & \multicolumn{1}{c}{\textbf{Old TPR}} & \multicolumn{1}{c}{\textbf{New TPR}} & \multicolumn{1}{c}{\textbf{Old TPR}} & \multicolumn{1}{c}{\textbf{New TPR}} & \multicolumn{1}{c}{\textbf{Old TPR}} & \multicolumn{1}{c}{\textbf{New TPR}} \\
\midrule
1\% & \multicolumn{1}{r}{99.9} & 6 & 5.4 & 83.3 & 1.2 & 99.8 & 2.8 & 55.8 & 0.3 & 99.4 & 4.5 & 94.15 & 3.8 & 99 \\
5\% & \multicolumn{1}{r}{100} & 18.8 & 16.6 & 93 & 5.7 & 99.8 & 10.1 & 87.4 & 2.1 & 99.9 & 16.7 & 97.88 & 14.7 & 99.4 \\
10\% & \multicolumn{1}{r}{100} & 28.2 & 25.3 & 96.5 & 13.1 & 99.8 & 17.9 & 94.1 & 4.3 & 99.9 & 29.2 & 98.39 & 25.9 & 99.8 \\
\bottomrule
\end{tabular}}
\caption{}
\end{subtable}

\hfill

\begin{subtable}[]{0.99\textwidth}
\scalebox{\scaleSize}{
\begin{tabular}{c|cc|cc|cc|cc|cc|cc|cc}
\toprule
\textbf{Watermarker} & \multicolumn{14}{c}{\textbf{\strong{KGW}}} \\
\midrule
\textbf{Translator} & \multicolumn{2}{c}{No Watermark} & \multicolumn{2}{c}{\textbf{\weak{KGW}}} & \multicolumn{2}{c}{\textbf{\strong{KGW}}} & \multicolumn{2}{c}{\textbf{\weak{PRW}}} & \multicolumn{2}{c}{\textbf{\strong{PRW}}} & \multicolumn{2}{c}{\textbf{\weak{SIR}}} & \multicolumn{2}{c}{\textbf{\strong{SIR}}} \\
\textbf{FPR} & \textbf{Gen TPR} & \multicolumn{1}{c}{\textbf{$D_W$}} & \multicolumn{1}{c}{\textbf{Old TPR}} & \multicolumn{1}{c}{\textbf{New TPR}} & \multicolumn{1}{c}{\textbf{Old TPR}} & \multicolumn{1}{c}{\textbf{New TPR}} & \multicolumn{1}{c}{\textbf{Old TPR}} & \multicolumn{1}{c}{\textbf{New TPR}} & \multicolumn{1}{c}{\textbf{Old TPR}} & \multicolumn{1}{c}{\textbf{New TPR}} & \multicolumn{1}{c}{\textbf{Old TPR}} & \multicolumn{1}{c}{\textbf{New TPR}} & \multicolumn{1}{c}{\textbf{Old TPR}} & \multicolumn{1}{c}{\textbf{New TPR}} \\
\midrule
1\% & \multicolumn{1}{r}{100} & 19.2 & 16.6 & 85.1 & 3.6 & 99.9 & 12 & 48 & 2.1 & 98.9 & 14.5 & 82.09 & 6.5 & 98.29 \\
5\% & \multicolumn{1}{r}{100} & 35 & 32.5 & 93.4 & 12.3 & 99.9 & 25.9 & 88.2 & 6.5 & 100 & 32.5 & 97.03 & 20.3 & 99.8 \\
10\% & \multicolumn{1}{r}{100} & 46.8 & 45 & 95.3 & 20.5 & 99.9 & 34.5 & 94.4 & 11.5 & 100 & 45.3 & 97.95 & 31.8 & 100 \\
\bottomrule
\end{tabular}}
\caption{}
\end{subtable}

\hfill
\hfill

\begin{subtable}[]{0.99\textwidth}
\scalebox{\scaleSize}{
\begin{tabular}{c|cc|cc|cc|cc|cc|cc|cc}
\toprule
\textbf{Watermarker} & \multicolumn{14}{c}{\textbf{\weak{PRW}}} \\
\midrule
\textbf{Translator} & \multicolumn{2}{c}{No Watermark} & \multicolumn{2}{c}{\textbf{\weak{KGW}}} & \multicolumn{2}{c}{\textbf{\strong{KGW}}} & \multicolumn{2}{c}{\textbf{\weak{PRW}}} & \multicolumn{2}{c}{\textbf{\strong{PRW}}} & \multicolumn{2}{c}{\textbf{\weak{SIR}}} & \multicolumn{2}{c}{\textbf{\strong{SIR}}} \\
\textbf{FPR} & \textbf{Gen TPR} & \multicolumn{1}{c}{\textbf{$D_W$}} & \multicolumn{1}{c}{\textbf{Old TPR}} & \multicolumn{1}{c}{\textbf{New TPR}} & \multicolumn{1}{c}{\textbf{Old TPR}} & \multicolumn{1}{c}{\textbf{New TPR}} & \multicolumn{1}{c}{\textbf{Old TPR}} & \multicolumn{1}{c}{\textbf{New TPR}} & \multicolumn{1}{c}{\textbf{Old TPR}} & \multicolumn{1}{c}{\textbf{New TPR}} & \multicolumn{1}{c}{\textbf{Old TPR}} & \multicolumn{1}{c}{\textbf{New TPR}} & \multicolumn{1}{c}{\textbf{Old TPR}} & \multicolumn{1}{c}{\textbf{New TPR}} \\
\midrule
1\% & \multicolumn{1}{r}{95.4} & 8.7 & 5.8 & 84.2 & 1.5 & 99.8 & 6.6 & 46.2 & 2.8 & 99.5 & 5.4 & 94.47 & 2.3 & 99.3 \\
5\% & \multicolumn{1}{r}{100} & 20.4 & 16.4 & 91.4 & 4.5 & 99.9 & 15.8 & 81.5 & 9.2 & 99.8 & 15.9 & 97.69 & 5.9 & 99.7 \\
10\% & \multicolumn{1}{r}{100} & 39.4 & 29.4 & 94.9 & 8.1 & 99.9 & 33.3 & 91.1 & 21.5 & 99.8 & 31.7 & 98.69 & 18.9 & 99.7 \\
\bottomrule
\end{tabular}}
\caption{}
\end{subtable}

\hfill

\begin{subtable}[]{0.99\textwidth}
\scalebox{\scaleSize}{
\begin{tabular}{c|cc|cc|cc|cc|cc|cc|cc}
\toprule
\textbf{Watermarker} & \multicolumn{14}{c}{\textbf{\strong{PRW}}} \\
\midrule
\textbf{Translator} & \multicolumn{2}{c}{No Watermark} & \multicolumn{2}{c}{\textbf{\weak{KGW}}} & \multicolumn{2}{c}{\textbf{\strong{KGW}}} & \multicolumn{2}{c}{\textbf{\weak{PRW}}} & \multicolumn{2}{c}{\textbf{\strong{PRW}}} & \multicolumn{2}{c}{\textbf{\weak{SIR}}} & \multicolumn{2}{c}{\textbf{\strong{SIR}}} \\
\midrule
\textbf{FPR} & \textbf{Gen TPR} & \multicolumn{1}{c}{\textbf{$D_W$}} & \multicolumn{1}{c}{\textbf{Old TPR}} & \multicolumn{1}{c}{\textbf{New TPR}} & \multicolumn{1}{c}{\textbf{Old TPR}} & \multicolumn{1}{c}{\textbf{New TPR}} & \multicolumn{1}{c}{\textbf{Old TPR}} & \multicolumn{1}{c}{\textbf{New TPR}} & \multicolumn{1}{c}{\textbf{Old TPR}} & \multicolumn{1}{c}{\textbf{New TPR}} & \multicolumn{1}{c}{\textbf{Old TPR}} & \multicolumn{1}{c}{\textbf{New TPR}} & \multicolumn{1}{c}{\textbf{Old TPR}} & \multicolumn{1}{c}{\textbf{New TPR}} \\
\midrule
1\% & \multicolumn{1}{r}{99.7} & 18.7 & 12.7 & 85.4 & 1.4 & 100 & 13.1 & 28.8 & 6.9 & 97.2 & 12.9 & 86.77 & 16.3 & 99.3 \\
5\% & \multicolumn{1}{r}{100} & 43.1 & 33.3 & 93 & 7.5 & 100 & 37.4 & 82.1 & 20.4 & 99.7 & 36.6 & 95.39 & 39 & 99.7 \\
10\% & \multicolumn{1}{r}{100} & 58.1 & 48.6 & 96.6 & 13.8 & 100 & 54 & 91.5 & 35 & 100 & 55.4 & 97.7 & 57.8 & 99.7 \\
\bottomrule
\end{tabular}}
\caption{}
\end{subtable}

\hfill

\begin{subtable}[]{0.99\textwidth}
\scalebox{\scaleSize}{
\begin{tabular}{c|cc|cc|cc|cc|cc|cc|cc}
\toprule
\textbf{Watermarker} & \multicolumn{14}{c}{\textbf{\weak{SIR}}} \\
\midrule
\textbf{Translator} & \multicolumn{2}{c}{No Watermark} & \multicolumn{2}{c}{\textbf{\weak{KGW}}} & \multicolumn{2}{c}{\textbf{\strong{KGW}}} & \multicolumn{2}{c}{\textbf{\weak{PRW}}} & \multicolumn{2}{c}{\textbf{\strong{PRW}}} & \multicolumn{2}{c}{\textbf{\weak{SIR}}} & \multicolumn{2}{c}{\textbf{\strong{SIR}}} \\
\midrule
\textbf{FPR} & \textbf{Gen TPR} & \multicolumn{1}{c}{\textbf{$D_W$}} & \multicolumn{1}{c}{\textbf{Old TPR}} & \multicolumn{1}{c}{\textbf{New TPR}} & \multicolumn{1}{c}{\textbf{Old TPR}} & \multicolumn{1}{c}{\textbf{New TPR}} & \multicolumn{1}{c}{\textbf{Old TPR}} & \multicolumn{1}{c}{\textbf{New TPR}} & \multicolumn{1}{c}{\textbf{Old TPR}} & \multicolumn{1}{c}{\textbf{New TPR}} & \multicolumn{1}{c}{\textbf{Old TPR}} & \multicolumn{1}{c}{\textbf{New TPR}} & \multicolumn{1}{c}{\textbf{Old TPR}} & \multicolumn{1}{c}{\textbf{New TPR}} \\
\midrule
1\% & \multicolumn{1}{r}{87.9} & 3.1 & 2.86 & 72 & 0.8 & 99.9 & 3.89 & 43.1 & 1.52 & 98.8 & - & - & - & - \\
5\% & \multicolumn{1}{r}{95.3} & 11.2 & 11.03 & 86.2 & 4.81 & 100 & 10.84 & 77.8 & 6.07 & 100 & - & - & - & - \\
10\% & \multicolumn{1}{r}{96.6} & 18.6 & 16.96 & 93.4 & 10.03 & 100 & 16.36 & 88.4 & 11.23 & 100 & - & - & - & - \\
\bottomrule
\end{tabular}}
\caption{}
\end{subtable}

\hfill

\begin{subtable}[]{0.99\textwidth}
\scalebox{\scaleSize}{
\begin{tabular}{c|cc|cc|cc|cc|cc|cc|cc}
\toprule
\textbf{Watermarker} & \multicolumn{14}{c}{\textbf{\strong{SIR}}} \\
\midrule
\textbf{Translator} & \multicolumn{2}{c}{No Watermark} & \multicolumn{2}{c}{\textbf{\weak{KGW}}} & \multicolumn{2}{c}{\textbf{\strong{KGW}}} & \multicolumn{2}{c}{\textbf{\weak{PRW}}} & \multicolumn{2}{c}{\textbf{\strong{PRW}}} & \multicolumn{2}{c}{\textbf{\weak{SIR}}} & \multicolumn{2}{c}{\textbf{\strong{SIR}}} \\
\midrule
\textbf{FPR} & \textbf{Gen TPR} & \multicolumn{1}{c}{\textbf{$D_W$}} & \multicolumn{1}{c}{\textbf{Old TPR}} & \multicolumn{1}{c}{\textbf{New TPR}} & \multicolumn{1}{c}{\textbf{Old TPR}} & \multicolumn{1}{c}{\textbf{New TPR}} & \multicolumn{1}{c}{\textbf{Old TPR}} & \multicolumn{1}{c}{\textbf{New TPR}} & \multicolumn{1}{c}{\textbf{Old TPR}} & \multicolumn{1}{c}{\textbf{New TPR}} & \multicolumn{1}{c}{\textbf{Old TPR}} & \multicolumn{1}{c}{\textbf{New TPR}} & \multicolumn{1}{c}{\textbf{Old TPR}} & \multicolumn{1}{c}{\textbf{New TPR}} \\
\midrule
1\% & \multicolumn{1}{r}{94.3} & 2.5 & 2.28 & 77.6 & 0.81 & 100 & 2.93 & 22.4 & 2.63 & 95.4 & - & - & - & - \\
5\% & \multicolumn{1}{r}{97.8} & 10.7 & 12.22 & 89.4 & 5.85 & 100 & 11.83 & 75.6 & 7.48 & 99.8 & - & - & - & - \\
10\% & \multicolumn{1}{r}{98.3} & 18.6 & 21.43 & 94.4 & 12.11 & 100 & 20.52 & 88.2 & 13.04 & 100 & - & - & - & - \\
\bottomrule
\end{tabular}}
\caption{}
\end{subtable}

\caption{TPR of the paraphrased text $T_P$ with dual watermarks when utilizing QWEN-7B as the base model of watermarker. FPR is set to $1\%$, $2\%$ \& $5\%$, respectively.}
\label{tab:qwen_paraphrase_results}
\end{sidewaystable*}

\begin{figure}[thb]
\small
\ttfamily
\begin{mdframed}

[INST] <<SYS>>
\newline
Assume you are a helpful assistant.
\newline
Your job is to translate the given text from LANGUAGE to LANGUAGE.
\newline
<</SYS>>
\newline
\texttt{\{INPUT\_TEXT\}} [/INST]
\newline
You're welcome! Here's a translated version of the original text:
\end{mdframed}
\caption{The back-translation prompt template for LLaMA-2 translator.}
\label{fig:prompts3}
\end{figure}

\begin{table*}
\newcommand{\cLen}{5cm}
\centering
\scalebox{0.55}{
\begin{tabular}{|c|p{\cLen}|p{\cLen}|p{\cLen}|p{\cLen}|p{\cLen}|}
\toprule
\diagbox{$W$}{$P$} & $\varnothing$ & $P'$ & \weak{KGW} & \weak{PRW} & \weak{SIR} \\
\midrule
\weak{KGW} & {in the future of our schools.
Voters approved \$700 million for school construction and improvement projects, a new building program and the sale of a new \$70 million bond. The new bond will be for projects to improve the air quality in our schools and make the buildings safer and more secure.
We are excited to share the following updates on the progress of the bond projects.
We have scheduled all projects through 2018 and are beginning to schedule some projects for the 2019 school year.
We are continuing to work with the District’s Capital Planning Committee to} & {
In the future of our schools, voters approved \$700 million for school construction and improvement projects, a new building program and the sale of a new \$70 million bond. The new bond will be for projects to improve the air quality in our schools and make the buildings safer and more secure. We are excited to share the following updates on the progress of the bond projects. We have scheduled all projects through 2018 and are beginning to schedule some projects for the 2019 school year. We are continuing to work with the District’s Capital Planning Committee to
} & {
in the future of our schools. Voters approved \$700 million for school construction and improvement projects, a new building program and the sale of a new \$70 million bond. The new bond will be for projects to improve the air quality in our schools and make the buildings safer and more secure. We are excited to share the following updates on the progress of the bond projects. We have scheduled all projects through 2018 and are beginning to schedule some projects for the 2019 school year. We are continuing to work with the District's Capital Planning Committee to
} & {
In the future, I hope that our school district will be able to make a lot of changes, like building new school buildings and making our schools more safe and secure.

Voters voted to let our school district sell \$70,000,000.00 of bonds for rebuilding and renovating our schools.

The reason why I decided to write this letter is because I want to make a change and help with our school district. We need to update the school buildings and make our schools safe and secure.
} & {
in our schools' futures. Approximately 700 million dollars were approved by voters for school construction and improvement projects, a new building program, and the sale of a new bond worth about 70 million dollars. The new bond will be used for initiatives that will make our schools' facilities safer and more secure as well as improve the air quality. We are thrilled to provide the following updates on the bond projects' status. All of the projects are scheduled through 2018, and we are starting to 
} \\
\midrule
\weak{PRW} & {
Jessie said, "I don't see her being a rude, bad person. What a lot of these women do when you see a lot of these women being rude, and you see a lot of these women not being respectful, and a lot of these women not being sweet, and a lot of these women being difficult,
} & {
Jessie said, "I don't see her being a rude, bad person. What a lot of these women do when you see a lot of these women being rude, and you see a lot of these women not being respectful, and a lot of these women not being sweet, and a lot of these women being difficult,
} & {
Jessie said, "I don't see her being a rude, bad person. What a lot of these women do when you see a lot of these women being rude, and you see a lot of these women being not respectful, and a lot of these women being not sweet, and a lot of these women being difficult, what you see when you see these women being difficult,
} & {
Jessie said, "I don't see her being a rude, bad person. What a lot of these women do when you see a lot of these women being rude, and you see a lot of these women being respectful, and a lot of these women not being respectful, and a lot of these women not being sweet, and a lot of these women being difficult,
} & {
Jessie said, I don't see her being a rude, bad person, but what I see is a lot of women being rude, and what I see is a lot of women not being respectful, and what I see is a lot of women not being sweet and difficult,
} \\
\midrule
\weak{SIR} & {
moving into agency work in 2008 and subsequently working on a freelance, interim and consultancy basis before re-joining Workhouse in 2018, working on a freelance contractor basis alongside a day job in a corporate agency environment and a project-based assignment in a digital transformation project in aerospace and defence, and a project-based assignment in aerospace and defence, before deciding to join on a permanent capacity and take on a client account and agency-wide responsibility for a leading client, which is currently in development and will be launched in
} & {
moving into agency work in 2008 and subsequently working on a freelance, interim and consultancy basis before re-joining Workhouse in 2018, working on a freelance contractor basis alongside a day job in a corporate agency environment and a project-based assignment in aerospace and defence, and a project-based assignment in aerospace and defence, before deciding to join on a permanent capacity and take on a client account and agency-wide responsibility for a leading client, which is currently in development and will be launched in 
} & {
moving into agency work in 2008 and subsequently working on a freelance, interim and consultancy basis before re-joining Workhouse in 2018, working on a freelance contractor basis alongside a day job in a corporate agency environment and a project-based assignment in aerospace and defence, and a project-based assignment in aerospace and defence, before deciding to join on
} & {
moving into agency work in 2008 and subsequently working on a freelance, interim and consultancy basis before re-joining Workhouse in 2018, working on a freelance contractor basis alongside a day job in a corporate agency environment and a project-based assignment in aerospace and defence, and a project-based assignment in aerospace and defence, before deciding to join on a permanent capacity and take on a client account and agency-wide responsibility for a leading client, which is currently in development and will be launched in 
} & N/A \\
\bottomrule
\end{tabular}
} 
\caption{Examples of watermarked texts under \textbf{weak} $W$, \textbf{weak} $P$ settings.}
\label{tab:example_weak2weak}
\end{table*}

\begin{table*}
\newcommand{\cLen}{5cm}
\centering
\scalebox{0.55}{
\begin{tabular}{|c|p{\cLen}|p{\cLen}|p{\cLen}|p{\cLen}|p{\cLen}|}
\toprule
\diagbox{$W$}{$P$} & $\varnothing$ & $P'$ & \weak{KGW} & \weak{PRW} & \weak{SIR} \\
\midrule
\weak{KGW} & {
the past, since the portable battery pack from ZMI provides mobile power so your mobile devices never run out again. You will never be out of the loop again. Whether at a business appointment in the city, a vacation in the country, a hike in the high alps - the ZMI portable batteries provide mobile power supply, so your mobile devices never let you down again. Choosing ZMI mobile batteries means investing money sensibly.
} & {
The ZMI portable power station provides mobile power so your mobile devices never run out again. You'll never be out of the loop again. Whether you're at a business meeting in the city, on vacation in the country, or hiking in the high alps - the ZMI portable batteries provide mobile power supply, so your mobile devices never let you down again. By choosing ZMI mobile batteries, you're investing money sensibly.
} & {
In the past, the portable power bank from ZMI was a lifesaver when your mobile devices were running out of power, so you never had to be out of the loop again. No matter where you were, whether in the city on a business trip, on vacation in the countryside, hiking in the high alps - the ZMI power bank always provided the power to keep your mobile devices up and running. When you buy
} & {
In the past, since the battery from ZMI provides mobile power, your mobile devices never run out again. You will never be out of the loop again.

Whether at a business appointment in the city, a vacation in the country, a hike in the high alps - the ZMI batteries provide mobile power supply,
so your mobile devices never let you down again.

Choosing ZMI mobile batteries means investing money sensibly.
} & {
In the past, your mobile devices never ran out of power again since ZMI's battery supplied mobile power. Never again will you be left behind.

ZMI batteries ensure that your mobile devices never fail you, whether you're on a work trip in the city, on holiday in the country, or hiking in the high alps.

Selecting ZMI mobile batteries is a wise financial decision.
} \\
\midrule
\weak{PRW} & {
ot seeks a full-time Director, Marketing and Communications. As a member and leader of the organization, the Director, Marketing and Communications will play a crucially important role and have the ability and skills needed for the organization and the community. With the recent completion and reopening of the historic theater, the community can look ahead and the new Director, Marketing and Communications can help define and execute a long-standing strategic marketing and communications strategy.
} & {
The community seeks a full-time Director, Marketing and Communications. As a member and leader of the organization, the Director, Marketing and Communications will play a crucially important role and have the ability and skills needed for the organization and the community. With the recent completion and reopening of the historic theater, the community can look ahead and the new Director, Marketing and Communications can help define and execute a long-standing strateg
} & {
A full-time Director of Marketing and Communications is needed for the community. The Director of Marketing and Communications will be a key member and leader of the organization, possessing the abilities and talents required for the community and the organization. The community can turn to the future now that the historic theater has been completed and reopened, and the new director of marketing and communications can assist in defining and carrying out a long-standing strategy.
} & {
Our organization seeks a new, forward-facing, and creative assistant with a varied skillset and an eagerness to jump headfirst and learn from our expert staff members, who have more combined years’ experience with our organization and its constituents’ needs and goals, and our overall vision—the most important, but sometimes difficult, task—which is that everyone, whether or not our organization is directly related or connected, is a member, and our organization’s members and supporters are our most important asset
} & {
Our organization is looking for a new, creative, forward-thinking assistant with a variety of skills and a willingness to dive right in and learn from our knowledgeable staff members who have more combined years of experience with our organization, the needs and goals of our constituents, and our overall vision—the most crucial, but occasionally challenging, task—which is that everyone is a member, whether or not our organization is directly related or connected, and that our members and supporters are our most valuable,
} \\
\midrule
\weak{SIR} & {
lending institutions and capital partners, as well as prospected new clientele and their professional advisors in their evaluative process of our client’ss financial information and capabilities, and its presentation of their financial and operational health and well- being, and their potential success and growth in their respective industry and region of operations and beyond, as well as their readiedness and ability in their ability and preparedness
} & {
Lending institutions and capital partners, as well as prospected new clientele and their professional advisors in their evaluative process of our client'ss financial information and capabilities, as well as their presentation of their financial and operational health and well- being, and their potential success and growth in their respective industry and region of operations and beyond, as well as their readiedness and ability in their ability and preparedness
} & {
during their audit and due diligence process, when our clients are looking for new lines and higher limits, as well as their presentation and representation of their financial and operations health and well, their potential and actual increase and overall value, as well as their readiness, suitability, and overall availability, we prospected new clients and their financial and business advisors
} & {
lending institutions and capital partners, as well as prospected new clients and their financial and business advisors, during their audit and due diligence process, when our clients are seeking new lines and higher limits, and their presentation and representation of their financial and operations health and well, and their potential and actual increase and overall value, and their preparedness and suitability and overall availabilty, 
} & N/A \\
\bottomrule
\end{tabular}
} 
\caption{Examples of watermarked texts under \textbf{strong} $W$, \textbf{strong} $P$ settings.}
\label{tab:example_strong2strong}
\end{table*}




\end{document}